\documentclass[9pt,twocolumn,a4paper]{article}

\usepackage{geometry}
\usepackage{amsmath}
\usepackage{amssymb}
\usepackage{graphicx}
\usepackage{foreign}

\usepackage[font=footnotesize]{caption}

\newif\ifarxiv
\newif\ifnotarxiv
\arxivtrue
\notarxivfalse

\newcommand{\myparagraph}[1]{\paragraph{#1.}}

\usepackage{palatino}

\usepackage{enumitem}
\usepackage{algorithmic}
\usepackage{algorithm}
\usepackage[dvipsnames]{xcolor}
\usepackage{xspace}
\usepackage{graphicx}
\usepackage{footnote}
\usepackage{hyperref}
\usepackage{multirow}
\usepackage{subfig}
\usepackage{enumitem}
\usepackage{comment}
\usepackage{amsmath}
\usepackage{amsfonts}
\usepackage{booktabs}

\newcommand{\faissfunction}[1]{{\fontfamily{lmtt}\color{blue}\selectfont #1}}

\newcommand{\metric}[1]{{\bf \color{OliveGreen} #1}}

\newcommand{\eg}{{e.g}.\@ }
\newcommand{\ie}{{i.e}.\@ }
\newcommand{\aka}{{a.k.a}.\@ }

\title{\sc The Faiss Library}
\date{}

\newcommand{\authplusaff}[2]{
\begin{minipage}{5cm}
\centering
{ #1} \\
\small{#2}
\vspace{4mm}
\end{minipage}
}

\author{
\begin{tabular}{ccc}
\authplusaff{Matthijs Douze}{FAIR, Meta} & 
\authplusaff{Alexandr Guzhva}{Zilliz} & 
\authplusaff{Chengqi Deng}{DeepSeek} \\
\authplusaff{Jeff Johnson}{FAIR, Meta} &
\authplusaff{Gergely Szilvasy}{FAIR, Meta} & 
\authplusaff{Pierre-Emmanuel Mazar{\'e}}{FAIR, Meta} \\
\authplusaff{Maria Lomeli}{FAIR, Meta} & 
\authplusaff{Lucas Hosseini}{Skip Labs} &
\authplusaff{Herv{\'e} J{\'e}gou}{FAIR, Meta} \\
\end{tabular}
}

\begin{document}

\maketitle

\begin{comment}
{\bf 
To the co-authors:
Please add comments with your $\backslash$name macro, not overleaf comments that are not visible when using the git interface.

We plan to have a good draft to distribute internally by Dec 22, publish the paper on ArXiV in January, and submit to a journal at the same moment.

Thanks!
}
\end{comment}

\begin{abstract}
Vector databases typically manage large collections of embedding vectors. 
As AI applications are growing rapidly, the number of embeddings that need to be stored and indexed is increasing. 
The Faiss library is dedicated to vector similarity search, a core functionality of vector databases.
Faiss is a toolkit of indexing methods and related primitives used to search, cluster, compress and transform vectors. 
This paper describes the trade-offs in vector search and the design principles of Faiss in terms of structure, approach to optimization and interfacing. 
We benchmark key features of the library and discuss a few selected use cases to highlight its broad applicability.

\end{abstract}

\section{Introduction}
\label{sec:intro}

\ifarxiv 
The 
\else 
\IEEEPARstart{T}{he}
\fi 
emergence of deep learning has induced a shift in how complex data is stored and searched, noticeably by the development of \emph{embeddings}. 
Embeddings are vector representations, typically produced by a neural network, that map (embed) the input media item into a vector space, where the locality encodes the semantics of the input.
Embeddings are extracted from various forms of media: words~\cite{mikolov2013efficient,bojanowski2017enriching}, text~\cite{devlin2018bert,izacard2021contriever}, images~\cite{caron2021emerging,pizzi2022self}, users and items for recommendation~\cite{paterek2007improving}.
They can even encode object relations, for instance multi-modal text-image or text-audio relations~\cite{duquenne2023sentence,radford2021learning}. 

Embeddings are employed as an intermediate representation for further processing, \eg self-supervised image embeddings are input to shallow supervised image classifiers~\cite{caron2018deep,caron2021emerging}.
They are also leveraged as a pretext task for self-supervision~\cite{chen2020simclr}.
In fact, embeddings are a compact intermediate representation that can be re-used for several purposes.

In this paper, we consider embeddings used directly to compare media items. 
The embedding extractor is designed so that the distance between embeddings reflects the similarity between their corresponding media. 
As a result, conducting neighborhood search in this vector space offers a direct implementation of similarity search between media items. 

Similarity search is also popular for tasks where end-to-end learning would not be cost-efficient. 
For example, a k-nearest-neighbor classifier is more efficient to upgrade with new training samples than a classification neural net. 
This explains why the usage of industrial database management systems (DBMS), that offer a vector storage and search functionality, has increased in the last years. 
These DBMS are at the junction of traditional databases and Approximate Nearest Neighbor Search (ANNS) algorithms. Until recently, the latter were mostly considered for specific use-cases or in research. 

From a practical perspective, the embedding extraction and the vector search algorithm are bound by an ``embedding contract'' on the embedding distance: 
\begin{itemize}
\item The embedding extractor, typically a neural network in modern systems, is trained so that distances between embeddings are aligned with the task to perform. 
\item The vector index performs neighbor search among the embedding vectors as accurately as possible w.r.t. exact search results given the agreed distance metric.
\end{itemize}

{\bf Faiss} is a library for ANNS. 
The core library is a collection of C++ source files without external dependencies.
Faiss also provides a comprehensive Python wrapper for its C++ core. 
It is designed to be used both from simple scripts and as a building block of a DBMS. 
In contrast with other libraries that focus on a single indexing method, 
Faiss is a toolbox that contains a variety of indexing methods that commonly involve a chain of components (preprocessing, compression, non-exhaustive search, etc.). 
In this paper, we show that there exists a choice between a dozen index types, and the optimal one usually depends on the problem's constraints. 

To summarize what Faiss is \emph{not}: 
Faiss does not extract features --  it only indexes embeddings that have been extracted by a different mechanism;
Faiss is not a service -- it only provides functions that are run as part of the calling process on the local machine; 
Faiss is not a database -- it does not provide concurrent write access, load balancing, sharding, transaction management or query optimization.
The scope of the library is intentionally limited to 
ANNS algorithmic implementation. 

The basic structure of Faiss is an \emph{index} that can have multiple implementations described in this paper. 
An index can store a number of \emph{database vectors} that are progressively added to it. 
At search time, a \emph{query vector} is submitted to the index. 
The index returns the database vector that is closest to the query vector w.r.t. the Euclidean distance. 
There are many variants of this functionality: 
instead of just the nearest neighbor, $k$ nearest neighbors are returned; 
instead of a fixed number of neighbors, only the vectors within a certain range are returned; 
batches of vectors can be searched in parallel;
other metrics besides the Euclidean distance are supported; 
the search can use either CPUs or GPUs.

Since its open-source release in 2017, Faiss has emerged as one of the most popular vector search libraries, boasting 37k GitHub stars and more than 5200 citations of its GPU implementation paper~\cite{johnson2019billion}. 
The Faiss packages have been downloaded 6M times. 
Major vector database companies, such as Zilliz and Pinecone, either rely on Faiss as their core engine or have reimplemented Faiss algorithms.

This paper exposes the design principles of Faiss. 
A similarity search library has to trade off between different constraints (Section~\ref{sec:tradeoffs}) using two main tools: vector compression (Section~\ref{sec:compression}) and non-exhaustive search (Section~\ref{sec:pruning}). 
We also review a few applications of Faiss for trillion-scale indexing, text retrieval, data mining, and content moderation (Section~\ref{sec:applications}). 
The appendix discusses how Faiss is structured and engineered to be flexible and usable from other tools. %
Throughout the paper, we refer to functions or classes in the Faiss codebase\footnote{\url{https://github.com/facebookresearch/faiss}} as well as the documentation\footnote{\url{https://faiss.ai/}} using \faissfunction{this specific style}.

\section{Related work}
\label{sec:related}

\myparagraph{Indexing methods}
Over the past decade, a steady stream of papers about indexing methods has been published. Faiss encompasses a broad range of algorithms, catering to a diverse spectrum of use cases. 

One of the most popular approaches in the industry is to employ Locality Sensitive Hashing (LSH) as a way to compress embeddings into compact codes. 
In particular, the cosine sketch \cite{charikar2002similarity} produces binary vectors such that the Hamming distance is an estimator of the cosine similarity between the original embeddings. 
The compactness of these sketches enables storing and searching very large databases of media content~\cite{lv2004image}, without the requirement to store the original embeddings. We refer the reader to the early survey by~\cite{wang2015learning} for research on binary codes. 

Since the original work introducing product quantization search~\cite{jegou2010product}, quantization-based ANN has emerged as a powerful alternative to binary codes~\cite{wang2017survey}. We refer the reader to the survey by~\cite{matsui2018survey} that discusses numerous research works related to quantization-based compact codes. 
 
The term LSH can also refer to indexing with multiple partitions, such as E2LSH~\cite{datar2004locality}. However, we do not consider this scenario as the performance is generally inferior to that of learnt partitions~\cite{pauleve2010locality}. 
Early data-aware methods that proved successful on large datasets include multiple partitions based on kd-tree or hierarchical k-means~\cite{muja2014scalable}. 
They are often combined with compressed-domain representation and are especially appropriate for very large-scale settings~\cite{jegou2008hamming,jegou2010product}.

After the introduction of NN-descent~\cite{dong2011efficient}, graph-based ANN algorithms have emerged as a viable alternative to space partitioning methods. Notably, HNSW, currently the most popular indexing method~\cite{malkov2018efficient} for medium-sized dataset, has been implemented in HNSWlib.

\myparagraph{Software packages}
Most of the research works on vector search have been open-sourced, and some of these evolved in relatively comprehensive software packages for vector search. 
For instance, FLANN includes several index types and a distributed implementation described extensively in~\cite{muja2014scalable}.
The first implementation of product quantization relied on the Yael library~\cite{douze2014yael}, that already had a few of the Faiss principles: optimized primitives for clustering methods (GMM and k-means), scripting language interface (Matlab and Python) and benchmarking operators. 
NMSlib, a package originally designed for text retrieval, was the first to include HNSW~\cite{boytsov2016off}. It also provides several index types.
The HNSWlib library later became the reference implementation of HNSW \cite{malkov2018efficient}.
Google's SCANN library is a thoroughly optimized implementation of IVFPQ~\cite{jegou2010product} on SIMD and includes several index variants for various database scales. 
SCANN was open-sourced but the accompanying paper~\cite{guo2020accelerating} omits to mention the engineering optimizations that underpin the library's remarkable speed. 
DiskANN~\cite{subramanya2019diskann} is Microsoft's foundational graph-based vector search library. 
It was originally built to leverage hybrid RAM/flash memory, but now  offers a RAM-only version and was later extended to perform efficient updates~\cite{singh2021freshdiskann}, out-of-distribution search~\cite{jaiswal2022ooddiskann} and 
filtered search~\cite{gollapudi2023filtered}.

Faiss was open-sourced concurrently with a publication~\cite{johnson2019billion} that details the GPU implementation of several index types.
The present paper complements this previous work by describing the library as a whole.

Concurrently, various software libraries from the database world were extended or developed to do vector search. 
Milvus~\cite{wang2021milvus} uses its Knowhere library, which relies on Faiss as one of its core engines. 
Pinecone~\cite{bruch2023approximate} initially relied on Faiss, although the engine was later rewritten in Rust. 
Weaviate~\cite{vanluijt2020bringing} is a composite retrieval engine that includes vector search among other methods.

\myparagraph{Theoretical guarantees}
Several approximate algorithms were inspired by the Johnson-Lindenstrauss lemma,  which stipulates that one can embed a set of high-dimensional points into a lower-dimensional space while almost preserving the distances. 
For instance, under certain assumptions about the data distribution, LSH algorithms based on random partitioning (e.g. projection) provide statistical guarantees for the range search problem. %
As this objective is only a proxy to the problem of nearest neighbor search, there are no formal guarantees w.r.t. any metrics such as nearest-neighbor recall. 
In practice, algorithms based on random partitioning are significantly outperformed by data-aware partitioning~\cite{pauleve2010locality,andoni2015optimal}. 
Similarly, graph-based algorithms are difficult to compare from a theoretical perspective if the data distribution is unknown. 

Techniques based on vector compression, either with binary sketches or quantization, often correspond to estimators of the target distance. 
For some of these methods, the variance of the estimator can be computed in closed form~\cite{charikar2002similarity,jegou2010product}, characterizing the distance estimation error. 
Again, the quality of the estimator is only a proxy for the actual k-nearest-neighbor problem. 

Considering these factors and the importance of execution speed, the quality of ANN search is typically evaluated by experimental comparisons on publicly available benchmarks.

\myparagraph{Benchmarks and competitions}
The leading benchmark for million-scale datasets is ANN-benchmarks~\cite{aumuller2020ann}. It compares about 50 implementations of ANNS. 
The big-ANN~\cite{simhadri2022results} challenge introduced large-scale settings, with 6 datasets containing 1 billion vectors each. 
Faiss was used as a baseline for the challenge and multiple submissions were derived from it. 
The 2023 edition of the challenge is at a smaller scale (10M vectors) but introduces more complex tasks, including a filtered track where Faiss served as a baseline method~\cite{bigann23}.

\myparagraph{Datasets}
Early datasets are based on keypoint features like SIFT \cite{lowe2004distinctive} used in image matching. 
We use BIGANN~\cite{jegou2011searching}, a dataset of 128-dimensional SIFT features. %
Later, when global image descriptors produced by neural nets became popular, the Deep1B dataset was released~\cite{babenko2016efficient}, with 96-dimensional image features extracted with Google LeNet \cite{szegedy2015going}.
For this paper, we introduce a dataset of 768-dimensional Contriever text embeddings~\cite{izacard2021contriever} that are compared using inner product similarity. 
The embeddings are computed with English Wikipedia passages.
The higher dimension of these embeddings is representative of contemporary applications. 

Each dataset has 10k query vectors and 20M to 350M training vectors. 
We indicate the size of the database explicitly, for example ``Deep1M'' means the database contains the 1M first vectors of Deep1B. 
The training, database and query vectors are sampled randomly from the same distribution: we do not address out-of-distribution data~\cite{jaiswal2022ooddiskann,baranchuk2023dedrift} in this work.

\section{Performance axes of a vector search library}
\label{sec:tradeoffs}

\begin{table}[]
    \centering
    \caption{Common notations used throughout the paper.}
    \label{tab:notations}
    \begin{tabular}{l|l}
    \toprule
    Notation & Meaning \\
    \midrule
    $d$  & vector dimension \\
    $N$  & number of vectors \\
    $q\in \mathbb{R}^d$  & query vector \\
    $x_i\in \mathbb{R}^d$  & $i$\textsuperscript{th} database vector \\
    $k$  & number of requested results \\
    $\varepsilon \in \mathbb{R}^+$  & radius for range search \\
    $K$  & number of centroids for quantization \\
    $M$  & number of sub-quantizers \\
    \bottomrule
    \end{tabular}
\end{table}

Vector search is a well-defined, unambiguous operation. 
In its simplest formulation, given a set of database vectors $\{x_i, i=1..N\} \subset \mathbb{R}^d$ and a query vector $q\in \mathbb{R}^d$, it computes
\begin{equation}
    n = \underset{i=1..N}{\mathrm{argmin}} \| q - x_i \|. 
\label{eq:basenns}
\end{equation}
The minimum can be computed with a direct algorithm by iterating over all database vectors: this is \emph{brute force search}. 
A slightly more general and complex operation is to compute the $k$ nearest neighbors of $q$:
\begin{equation}
    (n_1, ..., n_k, *, ..., *) =
    \underset{i=1..N}{\mathrm{argsort}} 
    \| q - x_i \|, 
\label{eq:knns}
\end{equation}
where $\mathrm{argsort}$ returns the indices of the array to sort it by increasing distances and $*$ means that an output is ignored. 
This is what the \faissfunction{search} method of a Faiss index returns.
A related operation is to find all the elements that are within some $\varepsilon$ distance to the query: 
\begin{equation}
    R =
    \{ 
    n =1..N
    \mathrm{\ s.t.\ }
    \| q - x_n \| \le \varepsilon
    \},
\label{eq:rangenn}
\end{equation}
which is computed with the \faissfunction{range\_search} method.

\myparagraph{Distance measures}
\label{sec:metricequiv}
In the equations above, we leave the definition of the distance undefined. 
Faiss aims at supporting a range of distance metrics, but in order to avoid overloading the library, it leaves out  metrics that are equivalent up to a monotonous transformation of distances (like Euclidean vs. squared Euclidean) or up to a preprocessing of the input vectors (like Mahalanobis or cosine, see below). 
The full list of metrics is in Appendix~\ref{app:metrics}.

The most commonly used distances in Faiss are the L2 distance, the cosine similarity and the inner product similarity (for the latter two, the $\mathrm{argmin}$ should be replaced with an $\mathrm{argmax}$).
These measures have useful analytical properties: for example, they are invariant under $d$-dimensional rotations. 

In terms of applications, the Euclidean distance is the default for comparing vectors~\cite{lowe2004distinctive}. 
When vectors are obtained by metric learning, the cosine similarity is often used to avoid collapsing or exploding embeddings~\cite{Deng2019arcface}. 
Maximum inner product search (MIPS) is most often used for recommendation systems that require to compare user and item embeddings~\cite{paterek2007improving}. 

These measures can be made equivalent by preprocessing transformations on the query and/or the database vectors. 
Table~\ref{tab:metricEq} summarizes the preprocessing transformations mapping different measures.
To our knowledge, some of these were already identified~\cite{bachrach2014speeding,hong2019asymmetric}, while others are new.

Note that vectors transformed in this manner have a very anisotropic distribution~\cite{morozov2018non}: the additional dimension incurred for many transformations is not homogeneous w.r.t. other dimensions. This can make indexing more difficult, in particular using product or scalar quantization methods. See Section~\ref{sec:preprocessing} for mitigations.

\newcommand{\drawat}[3]{\makebox[0pt][l]{\raisebox{#2}{\hspace*{#1}#3}}}

\begin{table*}[h]
    \caption{
    One wants to compare a query vector $x$ and a database vector $y$ given a particular metric (rows). 
    The table indicates how to preprocess $(x, y) \longmapsto (x', y') $ so that an index with another metric (columns) returns the nearest neighbors for the source metric. 
    Some cases require adding 1 or 2 extra dimensions to the original vectors, as is denoted by the vector concatenation symbol $[.;.]$.
    The positive scalar parameters $\alpha$ and $\beta$ are arbitrary and chosen to avoid negative values under a square root. 
    }
    \label{tab:metricEq}

    \centering
\begin{tabular}{c|c|c|c}
\toprule
index metric $\rightarrow$ & L2 & IP & cos \\
wanted metric $\downarrow$ & & & \\
\midrule
L2 & identity &
$
\begin{array}{l}
x' = [x ; -\alpha/2] \\
y' = [ y ; \| y \|^2 / \alpha ]
\end{array}
$
&
$
\begin{array}{l}
x' = [x ; -\alpha/2; 0] \\
y' = [ \beta y ; \beta \| y \|^2 / \alpha; \\
\sqrt{1 - \beta^2\|y\|^2 - \beta^2\|y\|^4/\alpha^2} 
]
\end{array}
$
\\
\hline
IP & 
$
\begin{array}{l}
x' = [x ; 0] \\
y' = [ y ; \sqrt{\alpha^2 - \| y \|^2} ]
\end{array}
$
& identity & 
$
\begin{array}{l}
x' = [x ; 0] \\
y' = [\alpha y ; \sqrt{1-\|\alpha y \|^2} ]
\end{array}
$
\\
\hline
cos & 
$
\begin{array}{l}
x' = x / \|x\| \\
y' = y / \|y\|
\end{array}
$
& 
$
\begin{array}{l}
x' = x / \|x\| \\
y' = y / \|y\|
\end{array}
$
& identity
\\
\bottomrule
\end{tabular}
\end{table*}

\subsection{Brute force search}

Developing an efficient implementation of brute force search is not trivial~\cite{chern2022tpu,johnson2019billion}.
It requires (1) an efficient way of computing the distances and (2) an efficient way of keeping track of the $k$ smallest distances. 

Computing distances in Faiss is performed either by direct distance computations, or, when query vectors are provided in large enough batches, using a matrix multiplication decomposition~\cite[Equation 2]{johnson2019billion}.
The corresponding Faiss functions are exposed in \faissfunction{knn} and 
\faissfunction{knn\_gpu} for CPU and GPU, respectively.

Collecting the top-$k$ smallest distances is usually %
done via a binary heap on CPU~\cite[section 2.1]{douze2014yael} or a sorting network on GPU~\cite{johnson2019billion,ootomo2023cagra}.
For larger values of $k$, it is more efficient to use a reservoir: an unordered result buffer of size $k'>k$ that is resized to $k$ when it overflows. 

Faiss's \faissfunction{IndexFlat} implements brute force search.
However, for large datasets this approach becomes too slow.
In low dimensions, there are branch-and-bound methods that yield exact search results. 
However, in large dimensions they provide no speedup over brute force search~\cite{weber1998quantitative}.

Besides, for some applications, small variations in distances are not significant enough to distinguish application-level positive and negative items~\cite{szilvasy2024vectorsearchsmallradiuses}. 
In these cases, approximate nearest neighbor search (ANNS) becomes interesting.

\subsection{Metrics for Approximate Nearest Neighbor Search}

With ANNS, the user accepts imperfect results, which opens the door to a new solution design space. 
With exact search, the database is represented as a plain matrix. 
For ANNS, the database may be preprocessed into an indexing structure, more simply referred to as an \emph{index} in the following. 

\myparagraph{Accuracy metrics}
In ANNS the \metric{accuracy}\footnote{Metrics involved in a tradeoff are indicated in a \metric{specific font}.} is measured as a discrepancy with the exact search results from \eqref{eq:knns} and \eqref{eq:rangenn}.
Note that this is an intermediate goal:
the end-to-end accuracy depends on (1) how well the distance metric correlates with the item matching objective and (2) the quality of ANNS, which is what we measure here.

Accuracy of k-nearest neighbor search is generally evaluated with ``$n$-recall@$k$'', which is the fraction of the $n$ ground-truth nearest neighbors that are in the $k$ first search results (with $n\le k$). 
Most often $k$\,$=$\,$1$ or $n$\,$=$\,$k$ (in which case the measure is \aka ``intersection measure''). 
When $n$\,=\,$k$\,=\,$1$, the recall measure and intersection are the same, and the recall is called ``accuracy''. 
In some publications~\cite{jegou2010product}, recall@$k$ means 1-recall@$k$, while in others~\cite{pmlr-v176-simhadri22a} it corresponds to $k$-recall@$k$.

For range search, the exact search result is obtained by applying \eqref{eq:rangenn}, using threshold $\varepsilon$. 
To yield result list $\widehat{R}$, the approximate search uses a (possibly different) threshold $\varepsilon'$. 
Thus, standard retrieval metrics can be computed: \metric{precision} $P=|R \cap \widehat{R}|/|\widehat{R}|$ and \metric{recall} $R=|R \cap \widehat{R}| / |R|$.
By sweeping $\varepsilon'$ from small to large, the result list $\widehat{R}$ increases,  producing a precision-recall curve. 
The area under the PR-curve is the \metric{mean average precision} score of range search~\cite{simhadri2022results}.
Setting $\varepsilon' \ne \varepsilon$ is relevant when approximate vector representations distort the distance metric (more on this in Section~\ref{sec:compression}). 

For vector encoder-decoder pairs, the standard metric is the mean squared error (MSE) between the original vector and the reconstructed vector~\cite{jegou2010product,babenko2014additive,huijben2024QINco}. 
For an encoder $C$ and a decoder $D$, the MSE is: 
\begin{equation}
    \mathrm{MSE} = 
    \mathbb{E}_x \big[ \| D(C(x)) - x \|_2^2 \big]. 
\end{equation}

\myparagraph{Resource metrics}
The other axes of the trade-off are related to computing resources. 
During search, the \metric{search time} and \metric{memory usage} are the main constraints. 
If compression is used, then, the memory usage can be smaller than what is required to store the original vectors. 

The index may need to store training data, which incurs a \metric{constant memory overhead} before any vector is added to the index.
The index can also add \metric{per-vector memory overhead} to the memory used to store each vector. 
This is the case for graph indexes, that need to store graph edges for each vector. 
The memory usage is more complex for settings with hybrid storage such as RAM + flash or GPU memory + RAM. 

The \metric{index building time} is also a resource constraint. 
It may be decomposed into a \metric{training time}, which is independent of the number of vectors added to the index, and the \metric{addition time per vector}.

In distributed settings or flash-backed storage, the relevant metric is the \metric{number of I/O operations} (IOPS), as each read operation fetches a whole page. 
Data layouts that minimize IOPs~\cite{subramanya2019diskann} are more efficient than small random accesses.
Another possibly limiting factor is the amount of \metric{extra memory} needed to pre-compute lookup tables used to speed up search.

\subsection{Tradeoffs}

Most often only a subset of metrics matter. 
For example, when a very large number of searches are performed on a fixed index, the \metric{index building time} does not matter. 
Or when the number of vectors is so small that the raw database fits in RAM multiple times, then the \metric{memory usage} does not matter. 
We refer to the metrics that we care about as the \emph{active constraints}. 
Note that \metric{accuracy} is always an active constraint because if it did not matter, returning random results would be sufficient (Faiss does actually provide an \faissfunction{IndexRandom} used in some benchmarking tasks).

In the following sections, we consider that the active constraints are  \metric{speed}, \metric{memory usage} and \metric{accuracy}. 
As such, we measure the \metric{speed} and \metric{accuracy} of several index types and hyperparameter settings under a fixed memory budget. 

\subsection{Exploring search-time settings}

\begin{figure}
    \centering
    \includegraphics[width=\linewidth]{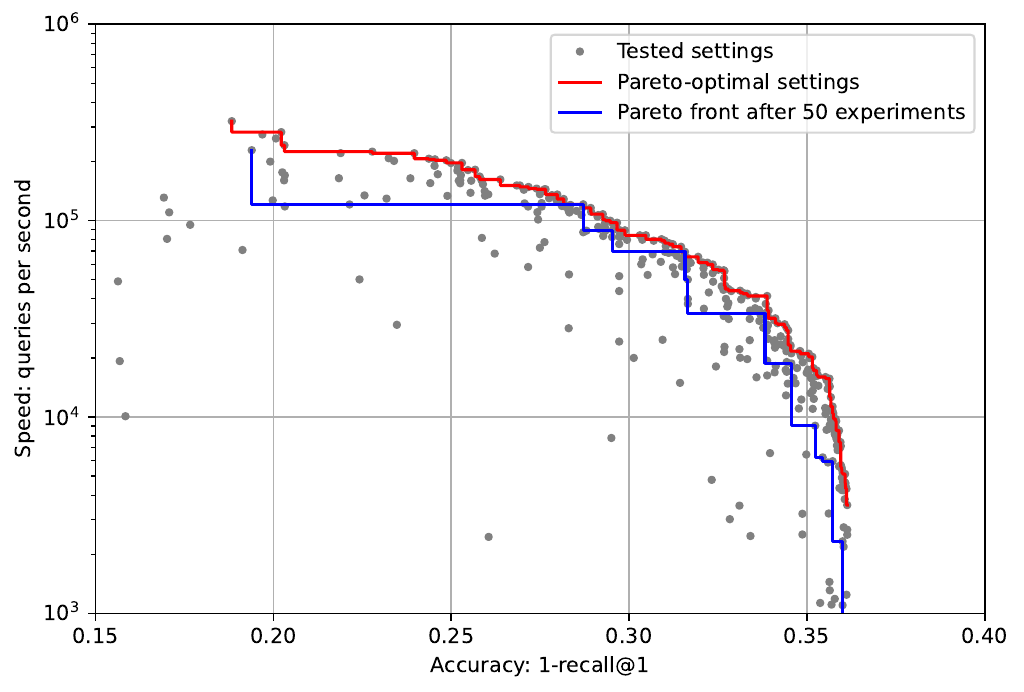}%
    \caption{
        Example of exploration of a parameter space with 3 parameters (an \faissfunction{IndexIVFPQ} with polysemous codes and HNSW coarse quantizer, running on the Deep100M dataset).
        The total number of configurations is 5808, but only 398 experiments are run. 
        We also show the set of operating points obtained with just 50 experiments. 
    }
    \label{fig:paretoexporation}
\end{figure}
For a fixed index, there are often one or several search-time hyperparameters that trade off speed with accuracy. 
For example, the \faissfunction{nprobe} hyperparameter for an \faissfunction{IndexIVF}, see Section~\ref{sec:pruning}. 
In general, we define hyperparameters as discrete scalar values such that when the value is higher, the \metric{speed} decreases and the \metric{accuracy} increases.
We can then keep only the Pareto-optimal settings, defined as settings that are the fastest for a given accuracy, or equivalently, that have the highest accuracy for a given time budget~\cite{sun2023automating}.

Exploring the Pareto-optimal frontier when there is a single hyper-parameter consists in sweeping over its values 
 with a certain level of granularity, and measuring the corresponding \metric{speed} and \metric{accuracy}. 

For multiple hyperparameters, the Pareto frontier can be recovered by exhaustively testing the Cartesian product of these parameters. 
However, the number of settings to test grows exponentially with the number of parameters.

\myparagraph{Pruning the parameter space}
one can leverage the monotonous nature of the hyperparameters for efficient pruning. 
We note a tuple of $n$ hyper-parameters $\pi = (p_1, ..., p_n) \in \mathcal{P} = \mathcal{P}_1\times ... \times \mathcal{P}_n$, and $\le$ a partial ordering on $\mathcal{P}$: $(p_1, .., p_n) \le (p'_1, .., p'_n) \Leftrightarrow \forall i, p_i \le p'_i$. Let $S(\pi)$ and $A(\pi)$ be the speed and accuracy obtained with this tuple of parameters. 
$\mathcal{P}^*\subset \mathcal{P}$ is the set of Pareto-optimal settings:
\begin{equation}
    \mathcal{P}^* = 
    \big\{ 
\pi \in \mathcal{P} |
\nexists \pi' \in \mathcal{P}
\mathrm{\ s.t.\ }
(S(\pi'),A(\pi')) > 
(S(\pi),A(\pi))
    \big\}. 
\end{equation}

Since the individual parameters have a monotonic effect on speed and accuracy, we have  
\begin{equation}
    \pi' \ge \pi
    \Rightarrow
    \left\{
    \begin{array}{c}
         S(\pi') \le S(\pi),  \\
         A(\pi') \ge A(\pi). \\
    \end{array}
    \right.
\end{equation}

Thus, if a subset $\widehat{\mathcal{P}} \subset \mathcal{P}$ of settings is already evaluated, the following upper bounds hold for a new setting $\pi \in \mathcal{P}$: 
\begin{eqnarray}
    S(\pi) &\le& \widehat{S}(\pi) = 
    \underset{\pi'\in \widehat{\mathcal{P}} \mathrm{\ s.t.\ } \pi' \le \pi}{\mathrm{Inf}}
    S(\pi'),  \\
    A(\pi) &\le& \widehat{A}(\pi) =
    \underset{\pi'\in \widehat{\mathcal{P}} \mathrm{\ s.t.\ } \pi' \ge \pi}{\mathrm{Inf}}
    A(\pi'). 
\end{eqnarray}

If any previous evaluation Pareto-dominates these bounds, the setting $\pi$ does not need to be evaluated: 
\begin{equation}
    \exists \pi' \in \widehat{\mathcal{P}}
    \mathrm{\ s.t.\ }
(S(\pi'),A(\pi')) > 
(\widehat{S}(\pi),\widehat{A}(\pi))
    \Rightarrow
\pi \notin \mathcal{P}^*. 
\label{eq:suboptimal}
\end{equation}

In practice, we evaluate settings from $\mathcal{P}$ in a random order. 
The pruning becomes more and more effective throughout the process. 
It is also more effective when the number of parameters is larger. 
Figure~\ref{fig:paretoexporation} shows an example with $|\mathcal{P}|=5808$ combined parameter settings. The pruning from \eqref{eq:suboptimal} reduces this to 398 experiments, out of which $| \mathcal{P}^*|=87$ are optimal. 
The Faiss \faissfunction{OperatingPoints} object implements this pruning.

\subsection{Refining (\faissfunction{IndexRefine})}
\label{sec:refine}

One can combine a fast but inaccurate index with a slower, more accurate search.~\cite{jegou2011searching,subramanya2019diskann,guo2020accelerating}. 
This is done by querying the fast index to retrieve a shortlist of results. 
The more accurate search then computes more accurate results only for the shortlist. 
This requires the accurate index to allow efficient random access to database vectors. 
Some implementations use a slower storage (e.g. flash) for the second index~\cite{subramanya2019diskann,sun2023soar}.

For the first-level index, the relevant accuracy metric is the recall at a rank equal to the shortlist size.
Thus, 1-recall@1000 can be a relevant metric, even if the end application does not use the 1000\textsuperscript{th} neighbor. 

Several methods based on this refining principle do not use two separate indexes. 
Instead, they use two ways of interpreting the same compressed vectors: a fast and inaccurate decoding and a slower but more accurate decoding~\cite{douze2016polysemous,douze2018link,Morozov2019UnsupervisedSearch,Amara2022NearestPerspective,huijben2024QINco} are based on this principle. 
The polysemous codes method~\cite{douze2016polysemous} is implemented in Faiss's \faissfunction{IndexIVFPQ}.

\section{Compression levels}
\label{sec:compression}

Faiss supports various vector codecs: these are methods to compress vectors so that they take up less memory.
A compression method $C:\mathbb{R}^d \rightarrow \{1,...,K\} $,  \aka a quantizer, converts a continuous multi-dimensional vector to an integer. 
This integer is equivalent to a bit string of code size $\lceil \log_2 K\rceil$. 
The decoder $D: \{1,...,K\} \rightarrow \mathbb{R}^d$ reconstructs an approximation of the vector from the integer.
The decoder can only reconstruct a finite number, $K$, of distinct vectors. 

The search of \eqref{eq:basenns} becomes approximate: 
\begin{equation}
    n = \underset{i=1..N}{\mathrm{argmin}} \| q - D(C(x_i)) \|
    = \underset{i=1..N}{\mathrm{argmin}} \| q - D(C_i) \|, 
\label{eq:adc}
\end{equation}
where the codes $C_i = C(x_i)$ are precomputed and stored in the index.
This is the asymmetric distance computation (ADC)~\cite{jegou2010product}. 
The symmetric distance computation (SDC) corresponds to the case when the query vector is also compressed:
\begin{equation}
    n = \underset{i=1..N}{\mathrm{argmin}} \| D(C(q)) - D(C_i) \|. 
\label{eq:sdc}
\end{equation}

Most Faiss indexes perform ADC as it is more accurate: no accuracy is lost on the query vectors. 
SDC is useful when there is also a storage constraint on the queries or for indexes where SDC is faster to compute than ADC.
The naive computation of \eqref{eq:adc} decompresses the vectors, which has an impact on speed. 
In most cases, the distance can be computed in the compressed domain. %

\subsection{The vector codecs}

\myparagraph{The k-means vector quantizer (\faissfunction{Kmeans})}
The ideal vector quantizer minimizes the MSE between the original and the decompressed vectors. 
This is formalized in the Lloyd necessary conditions for the optimality of a quantizer~\cite{lloyd1982least}.

The k-means algorithm directly implements these conditions. %
The $K$ centroids of k-means are an explicit enumeration of all possible vectors that can be reconstructed.

The k-means vector quantizer is very \metric{accurate} but the \metric{memory usage} and \metric{encoding complexity} grow exponentially with the code size. 
Therefore, k-means is impractical to use beyond roughly 3-byte codes, corresponding to 16M centroids.

\myparagraph{Scalar quantizers}
Scalar quantizers encode each dimension of a vector independently. 

A very classical and simple scalar quantizer is LSH (\faissfunction{IndexLSH}), where each vector component is encoded in a single bit by comparing it to a threshold. 
The threshold can be fixed to 0 or trained.
Faiss supports efficient SDC search of binary vectors via the \faissfunction{IndexBinary} objects, see Section~\ref{sec:binaryindexes}.

The \faissfunction{ScalarQuantizer} also supports uniform quantizers that encode a vector component into 8, 6 or 4 bits -- referred to as \verb|SQ8|, \verb|SQ6|, \verb|SQ4|. 
A scale and offset determine which values are reconstructed. 
They can be set separately for each dimension on the whole vector. 
The \faissfunction{IndexRowwiseMinMax} stores vectors with per-vector normalizing coefficients.
Lower-precision 16-bit floating point representations are also considered as scalar quantizers, \verb|SQfp16| and \verb|SQbf16|.

\myparagraph{Multi-codebook quantizers}
Faiss contains several multi-codebook quantization (MCQ) options. 
They are built from $M$ vector quantizers that can reconstruct $K$ distinct values each. 
The codes produced by these methods are of the form $(c_1, ..., c_M)\in\{1,...,K\}^M$, \ie each code indexes one of the quantizers.
The number of reconstructed vectors is $K^M$ and the code size is thus $M\lceil \log_2(K) \rceil$.

The product quantizer (\faissfunction{ProductQuantizer}, also noted PQ) is a simple MCQ that splits the input vector into $M$ sub-vectors and quantizes them separately~\cite{jegou2010product} with a k-means quantizer. 
At reconstruction time, the individual reconstructions are concatenated to produce the final code. 
In the following, we will use the notation \verb|PQ6x10| for a product quantizer with 6 sub-vectors each encoded in 10 bits ($M=6$, $K=2^{10}$).

Additive quantizers are a family of MCQ where the reconstructions from sub-quantizers are summed up together. 
Finding the optimal encoding for a vector given the codebooks is NP-hard~\cite{babenko2014additive}, so, in practice, additive quantizers use heuristics to find near-optimal codes. 

Faiss supports two types of additive quantizers. 
The residual quantizer (\faissfunction{ResidualQuantizer}) proceeds sequentially, by encoding the difference (residual) of the vector to encode and the one that is reconstructed by the previous sub-quantizers~\cite{chen2010approximate}.
The local search quantizer (\faissfunction{LocalSearchQuantizer}) starts from a sub-optimal encoding of the vector and locally explores neighbording codes in a simulated annealing process~\cite{martinez2016revisiting,martinez2018lsq++}.
We use notations \verb|LSQ6x10| and \verb|RQ6x10| to refer to additive quantizers with 6 codebooks of size $2^{10}$.

Faiss also supports a combination of PQ and additive quantizer, \faissfunction{ProductResidualQuantizer}. %
In that case, the vector is split in sub-vectors that are encoded independently with additive quantizers~\cite{babenko2015tree}.
The codes from the sub-quantizers are concatenated. 
We use the notation \verb|PRQ2x6x10| to indicate that vectors are split in 2 and encoded independently with \verb|RQ6x10|, yielding a total of 12 codebooks of size $2^{10}$.

\myparagraph{Hierarchy of quantizers}
Although this is not by design, there is a strict ordering between the quantizers described before. 
This means that quantizer $i+1$ can have the same set of reproduction values as quantizer $i$: it is more flexible and more data adaptive. 
The hierarchy of quantizers is: 
\begin{enumerate}
    \item 
    the binary representation with bits +1 and -1 can be represented as a scalar quantizer with 1 bit per component;
    \item
    the scalar quantizer is a product quantizer with 1 dimension per sub-vector and uniform per-dimension quantizer;
    \item 
    the product quantizer is a product-additive quantizer where the additive quantizer has a single level;
    \item 
    the product additive quantizer is an
    additive quantizer where within each codebook all components outside one sub-vector are set to 0~\cite{babenko2014additive};
    \item 
    the additive quantizer is the general case where the codebook entries correspond to all possible reconstructions obtained by adding elements from the subquantizers.
\end{enumerate}

The implications of this hierarchy are 
(1) the degrees of freedom for the reproduction values of quantizer $i+1$ are larger than for $i$, so it is more \metric{accurate} 
(2) quantizer $i+1$ has a higher capacity so it consumes more resources in terms of \metric{training time} and \metric{storage overhead} than $i$. 
In practice, the product quantizer often offers a good trade-off, which explains its wide adoption.
The corresponding Faiss \faissfunction{Quantizer} objects are listed in Appendix~\ref{app:quantizers}.

\begin{figure*}
    \centering
    \includegraphics[width=\columnwidth]{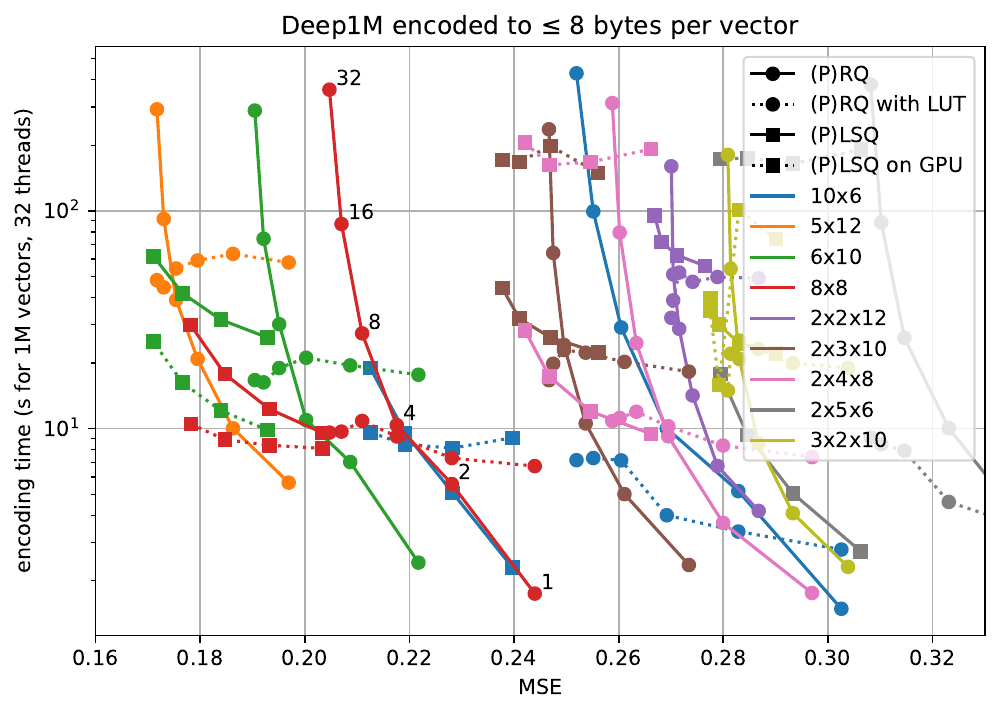}
    \includegraphics[width=\columnwidth]{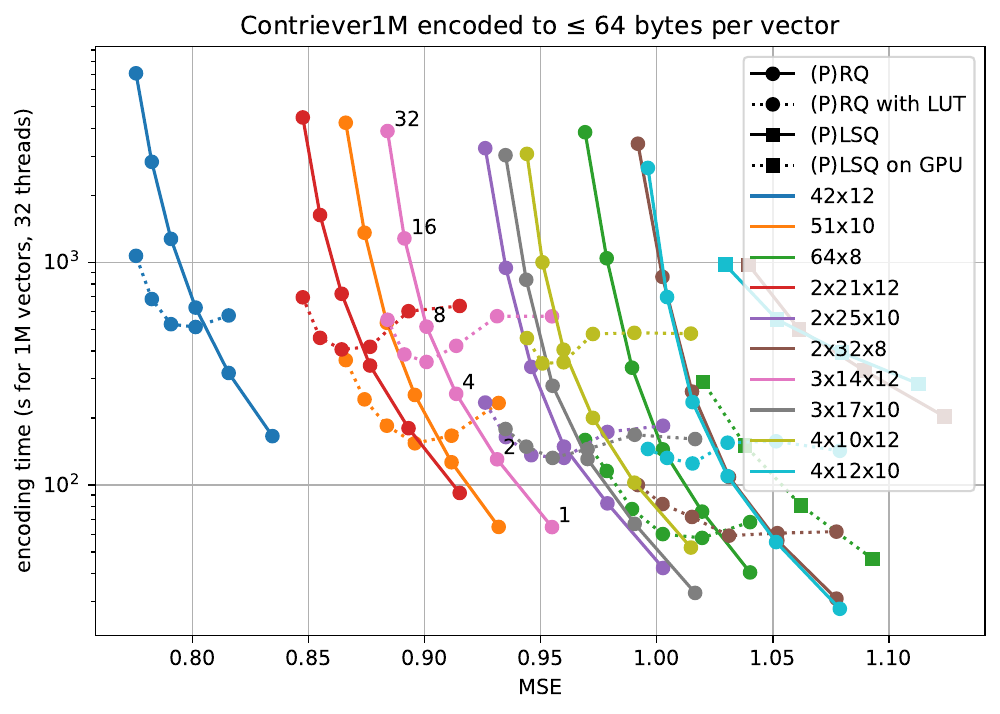}
    \caption{Comparison of additive quantizers in terms of \metric{encoding time} vs. \metric{accuracy} (MSE).
    Lower values are better for both. 
    We consider two different regimes: 
    Deep1M (low-dimensional) to 8-bytes codes and Contriever1M (high dimensional) to 64-byte codes.
    For some RQ variants, we indicate the beam size setting at which that trade-off was obtained.
    }
    \label{fig:aqoptions}
\end{figure*}

\subsection{Vector preprocessing}
\label{sec:preprocessing}

Applying transformations to input vectors before encoding can enhance the effectiveness of certain quantizers. 
In particular, $d$-dimensional rotations are commonly used, as they preserve comparison metrics like cosine, L2 and inner product. 

Scalar quantizers assign the same number of bits per vector component. 
However, for distance comparisons, if specific vector components have a higher variance, they have more impact on the distances. 
In other works, a variable number of bits are assigned per component~\cite{sandhawalia2010searching}. 
However, it is simpler to apply a random rotation to the input vectors, which in Faiss can be done with a \faissfunction{RandomRotationMatrix}. 
The random rotation spreads the variance over all the dimensions without changing the measured distances. 

An important transform is the Principal Component Analysis (PCA), that reduces the number of dimensions $d$ of the input vectors to a user-specified $d'$. 
This operation (\faissfunction{PCAMatrix}) is the orthogonal linear mapping that best preserves the variance of the input distribution.
It is often beneficial to apply PCA to large input vectors before quantizing them as k-means quantizers are more likely to ``fall'' in local minima in high-dimensional spaces~\cite{liu2015improved,jegou2011aggregating}.

The OPQ transformation~\cite{ge2013optimized} is a rotation of the input space that decorrelates the distribution of each sub-vector of a product quantizer\footnote{In Faiss terms, OPQ and ITQ are preprocessings. The actual quantization is performed by a subsequent product quantizer or binarization step.}. 
This makes PQ more accurate in the case where the variance of the data is concentrated on a few components.
The Faiss implementation \faissfunction{OPQMatrix} combines OPQ with a dimensionality reduction.
The ITQ transformation~\cite{gong2012iterative} similarly rotates the input space prior to binarization (\faissfunction{ITQMatrix}).

\subsection{Faiss additive quantization options}

Additive quantizers exist in two main variants: the residual quantizer and local search quantizer.
They are more complex than most quantizers because the \metric{index building time} must be taken into account. 
In fact, the accuracy of an additive quantizer of a certain size can always be increased at the cost of an increased encoding time (and training time).

Additive quantizers are based on $M$ codebooks $T_1,...T_M$ of size $K$. 
The decoding of code $C(x) = (c_1,...,c_M)$ is 
\begin{equation}
    x' = D(C(x)) = T_1[c_1] + ... + T_M[c_M]. 
\end{equation}

Thus, decoding is unambiguous. 
However, there is no practical way to encode vectors optimally, let alone train the codebooks. 
Enumerating all possible encodings is of exponential complexity in $M$. 

\myparagraph{The residual quantizer (RQ)}
RQ encodes a vector $x$ sequentially. 
At stage $m$, RQ picks the entry that best reconstructs the residual of $x$ w.r.t. the previous encoding steps: 
\begin{equation}
      c_m = 
      \underset{j=1..K}{\mathrm{argmin}} 
      \left\| \sum_{i=1}^{m-1} T_i[c_i] + T_m[j] - x\right\|^2. 
      \label{eq:rqencoding}
\end{equation}
This greedy approach tends to get trapped in local minima. 
As a mitigation, the encoder maintains a beam of  \faissfunction{max\_beam\_size} of possible codes and picks the best code at stage $M$. 
This parameter adjusts the trade-off between \metric{encoding time} and \metric{accuracy}.

To speed up the encoding, the norm of \eqref{eq:rqencoding} can  be decomposed into the sum of: 
\begin{itemize}
\item 
    $\| T_m[j] \|^2$  is precomputed and stored; 
\item 
    $\left\| \sum_{i=1}^{m-1} T_i[c_i] - x \right\|^2$ is the 
    encoding error of the previous step $m-1$; 
\item 
    $- 2 \langle T_m[j], x \rangle$  is computed on entry to the encoding (it is the only computation complexity that depends on $d$);
\item 
    $2 \sum_{\ell=1}^{m-1} \langle T_m[j], T_\ell[c_\ell] \rangle$ is also precomputed. 
\end{itemize}
This decomposition is used when \faissfunction{use\_beam\_LUT} is set. 
It is interesting only if $d$ is large and when $M$ is small because the storage and compute requirements of the last term grow quadratically with $M$.

\myparagraph{The local search quantizer (LSQ)}
At encoding time, LSQ starts from a suboptimal encoding  of the vector and proceeds with a simulated annealing optimization to refine the codes. 
At each optimization step, LSQ randomly flips codes and then uses Iterated Conditional Mode (ICM) to optimize the new encoding.
The number of optimization steps is set with \faissfunction{encode\_ils\_iters}. 
The LSQ codebooks are trained via an expectation-maximization procedure (similar to k-means).

\begin{figure*}
    \centering
    \begin{tabular}{c@{}c}
    Deep1M & Contriever1M \\
    \includegraphics[width=0.5\linewidth]{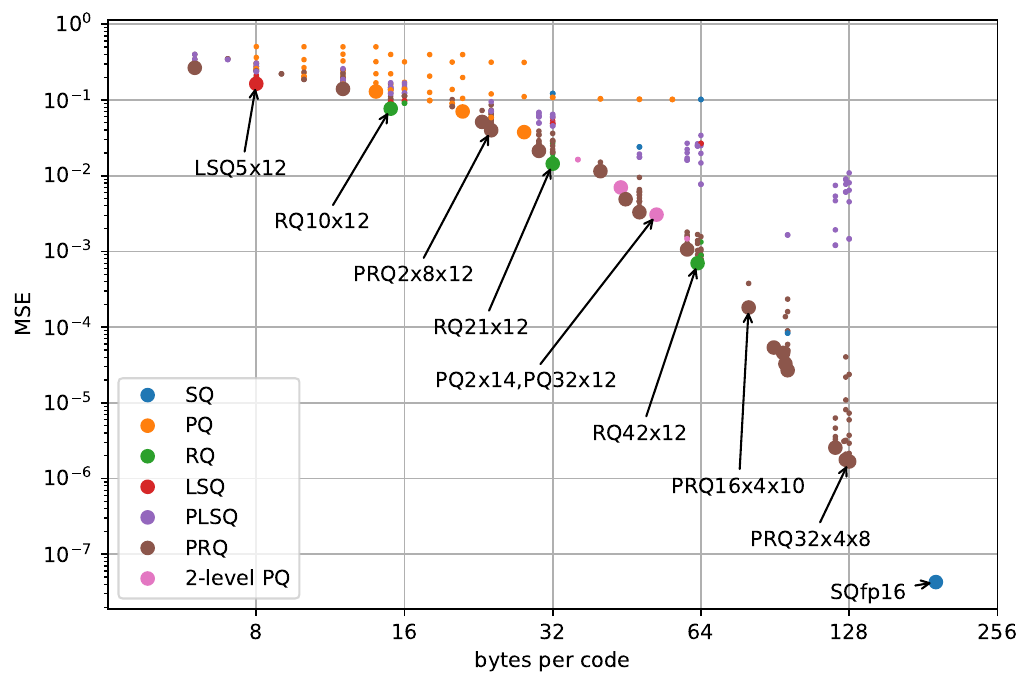} & 
    \includegraphics[width=0.5\linewidth]{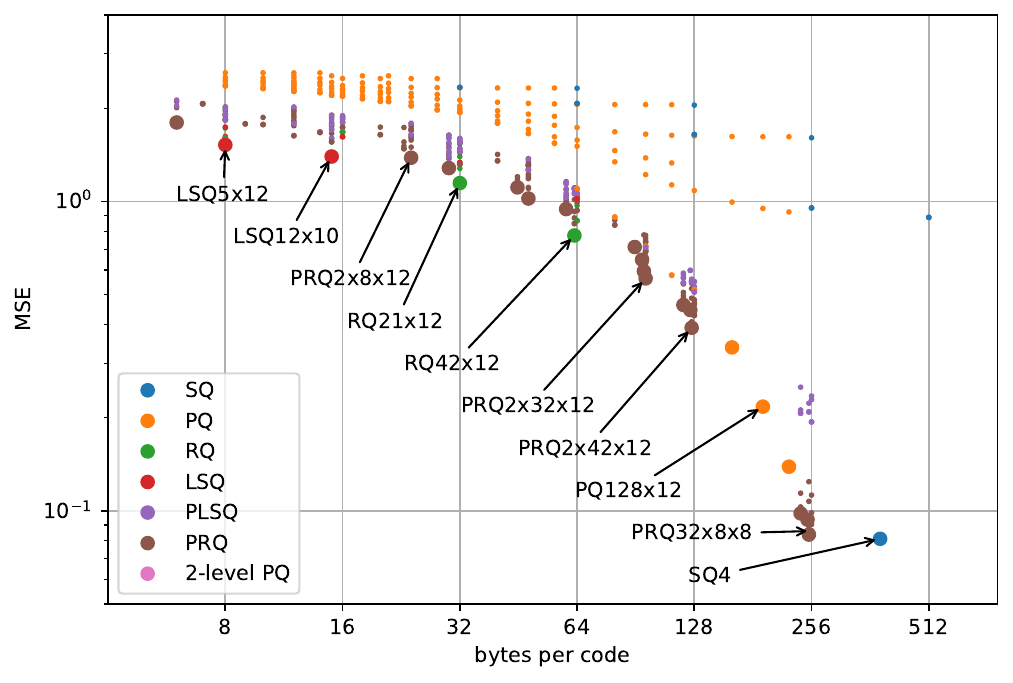} \\
    \end{tabular}
    \caption{
         \metric{Accuracy} vs. \metric{code size} trade-off for different codecs on the Deep1M and Contriever1M datasets.    
        We show Pareto-optimal variants with larger dots and indicate the quantizer in text for some of them. 
        Note that contriever vectors can be encoded to MSE=$2\cdot 10^{-4}$ in 768 bytes with \texttt{SQ8} (that setting is widely out-of-range for the plot).
    }
    \label{fig:codecs}
\end{figure*}

\myparagraph{Compressed-domain search}
The distances are computed without decompressing the stored vectors. 
It is acceptable to perform pre-computations on the query vector $q$ because the cost of these pre-computations is amortized over many query-to-code distance comparisons.

Additive quantizer inner products can be computed in the compressed domain: 
\begin{equation}
   \langle q, x' \rangle =
   \sum_{m=1}^M 
   \langle T_m[c_m], q \rangle =
   \sum_{m=1}^M 
   \mathrm{LUT}_m[c_m]. 
\label{eq:lutlookup}
\end{equation}
The lookup tables $\mathrm{LUT}_m$ are computed when a query vector comes in, similar to product quantizer search~\cite{jegou2010product}. 

This decomposition does not work to compute L2 distances. %
As a workaround, Faiss uses the decomposition~\cite{babenko2014additive}
\begin{equation}
\| q - x' \|^2 = \|q\|^2 + \|x'\|^2 - 2 \langle q, x' \rangle. 
\end{equation}
Thus, the term $\|x'\|^2$ must be available at search time. 
Using the \faissfunction{AdditiveQuantizer.search\_type} configuration, it can be appended in the stored code (\faissfunction{ST\_norm\_float32}), possibly compressed (\faissfunction{ST\_norm\_qint8}, \faissfunction{ST\_norm\_qint4},...).
It can also be computed on the fly (\faissfunction{ST\_norm\_from\_LUT}) with
\begin{equation}
    \|x'\|^2 = 
    2 \sum_{m=1}^M \sum_{\ell=1}^{m - 1} 
    \langle T_m[c_m], T_\ell[c_\ell] \rangle + 
    \sum_{m=1}^M  \| T_m[c_m] \|^2. 
\end{equation}
There, the norms and dot products are stored in the same lookup tables as the one used for beam search. 
Therefore, it trades off \metric{search time} for \metric{memory} overhead to store codes. 

Figure~\ref{fig:aqoptions} shows the trade-off between \metric{encoding time} and \metric{MSE}. 
Given a code size, it is more accurate to use a smaller number of sub-quantizers $M$ and a higher $K$. 
GPU encoding for LSQ does not help systematically. 
The LUT-based encoding of RQ is interesing for RQ/PRQ quantization when the beam size is larger.
In the 64-byte regime, we observe that LSQ is not competitive with RQ. 
PLSQ and PRQ progressively become more competitive for larger memory budgets.
They are also faster, since they operate on smaller vectors.

\subsection{Vector compression benchmark}

Figure~\ref{fig:codecs} shows the trade-off between \metric{code size} and \metric{accuracy} for many variants of the codecs. 
Additive quantizers are the best options for small code sizes. 
For larger code sizes, it is beneficial to independently encode several sub-vectors with product-additive quantizers. 
LSQ is more accurate than RQ for small codes, but does not scale well to longer codes. 
Note that product quantizers are a bit less accurate than additive quantizers, but given their low \metric{encoding time} they remain an attractive option. 
The scalar quantizers perform well for very long codes and are even faster. 
The 2-level PQ options are what an IVFPQ index uses as encoding: a first-level coarse quantizer and a second level refinement of the residual (more about this in Section~\ref{sec:residuals}).

\subsection{Binary indexes}
\label{sec:binaryindexes}

Binary quantization with symmetric distance computations is a pattern that has been commonly used~\cite{wang2015learning,Cao_2017_ICCV}. In this setup, distances are computed in the compressed domain as  Hamming distances. \eqref{eq:sdc} reduces to: 
\begin{equation}
    n = \underset{i=1..N}{\mathrm{argmin}}
    \| C(q) - C_i \|. 
\end{equation}
where $C(q), C_i \in \{0, 1\}^d$. 
Hamming distances are integers in $\{0..d\}$. Although they are crude approximations for continuous domain distances, they are fast to compute, do not require any specific context, and are easy to calibrate in practice. 

The \faissfunction{IndexBinary} indexes support addition and search directly from binary vectors. They offer a compact representation and leverage optimized instructions for distance computations.

The simplest \faissfunction{IndexBinaryFlat} index performs exhaustive search. 
Three options are offered for non-exhaus\-tive search:
\begin{itemize}
\item{} \faissfunction{IndexBinaryIVF} is a binary counterpart for the inver\-ted-list \faissfunction{IndexIVF} index described in \ref{invertedfiles}.
\item{} \faissfunction{IndexBinaryHNSW} is a binary counterpart for the hierarchical graph-based \faissfunction{IndexHNSW} index described in \ref{graphbased}.
\item{} \faissfunction{IndexBinaryHash} uses prefix vectors as hashes to cluster the database (rather than spheroids as with inverted lists), and searches only the clusters with closest prefixes.
\end{itemize}

Finally, the\faissfunction{IndexBinaryFromFloat} is provided for convenience. It wraps an arbitrary index and offers a binary vector interface for its operations.

\section{Non-exhaustive search}
\label{sec:pruning}

Non-exhaustive search is the cornerstone of fast search implementations for datasets larger than around $N$=10k vectors. 
In that case, the aim of the indexing method is to quickly focus on a subset of database vectors that are most likely to contain the search results. 

A method to do this is Locality Sensitive Hashing 
(LSH).
It amounts to projecting the vectors on a random direction~\cite{datar2004locality}. 
The offsets on that direction are then discretized into buckets where the database vectors are stored.
At search time, only the nearest buckets to the query vector's projection are visited. 
In practice, \emph{several} projection directions are needed to make it \metric{accurate}, at the cost of \metric{search time} and \metric{memory usage}. 
A fundamental drawback of this method is that it is not data-adaptive, although some improvements are possible~\cite{pauleve2010locality}.

An alternative way of pruning the search space is to use tree-based indexing. 
In that case, the dataset is stored in the leaves of a tree~\cite{muja2014scalable}.
When querying a vector, the search starts at the root node. 
At each internal node, the search descends into one of the child nodes depending on a decision rule. 
The decision rule depends on how the tree was built: 
for a KD-tree it is the position w.r.t. a hyperplane, for a hierarchical k-means, it is the proximity to a centroid. 

LSH and tree-based methods both aim to extend classical database search structures to vector search, because they have a favorable complexity (constant or logarithmic in $N$). 
However, these methods do not scale well for dimensions above 10. 

Faiss implements two non-exhaustive search approaches that operate at different \metric{memory} vs. \metric{speed} trade-offs: inverted file and graph-based. 

\subsection{Inverted files} \label{invertedfiles}

\def\nlist{{K_\mathrm{IVF}}}
\def\nprobe{{P_\mathrm{IVF}}}

IVF indexing is a technique that clusters the database vectors at indexing time.
This clustering uses a vector quantizer (the \emph{coarse quantizer}) that outputs $\nlist$  distinct indices (the \faissfunction{nlist} field of the \faissfunction{IndexIVF} object).
The coarse quantizer's $\nlist$ reproduction values are called \emph{centroids}. 
The vectors of each cluster (possibly compressed) are stored contiguously into  inverted lists, forming an inverted file (IVF).
At search time, only a subset of $\nprobe$  clusters are visited (\aka \faissfunction{nprobe}). 
The subset is formed by searching the $\nprobe$ nearest centroids, as in \eqref{eq:knns}.

\myparagraph{Setting the number of lists}
\label{sec:nlist}
The $\nlist$ parameter is central. 
In the simplest case, when $\nprobe$ is fixed, the coarse quantizer is exhaustive, the inverted lists contain uncompressed vectors, and the inverted lists are all the same size, then the number of distance computations is 
\begin{equation}
N_\mathrm{distances} = \nlist + \nprobe \times N / \nlist
\label{eq:centroidfactor}
\end{equation}
reaching a minimum when $\nlist = \sqrt{\nprobe N}$. 
This yields the usual recommendation to set $\nprobe$ proportional to $\sqrt{N}$.

In practice, this is just a rough approximation because 
(1) the $\nprobe$ has to increase with the number of lists in order to keep a fixed \metric{accuracy}
(2) the inverted lists sizes are not balanced 
(3) often the coarse quantizer is not exhaustive itself, so the quantization uses fewer than $\nlist$ distance computations, for example it is common to use a non-exhaustive HNSW index to perform the coarse quantization. 

The \emph{imbalance factor} is the relative variance of inverted list sizes~\cite{tavenard2011balancing}. 
At search time, if the inverted lists all have the same length, this factor is 1. 
If they are unbalanced, the expected number of distance computations is multiplied by this factor.

\begin{figure}
    \centering
    \includegraphics[width=\columnwidth]{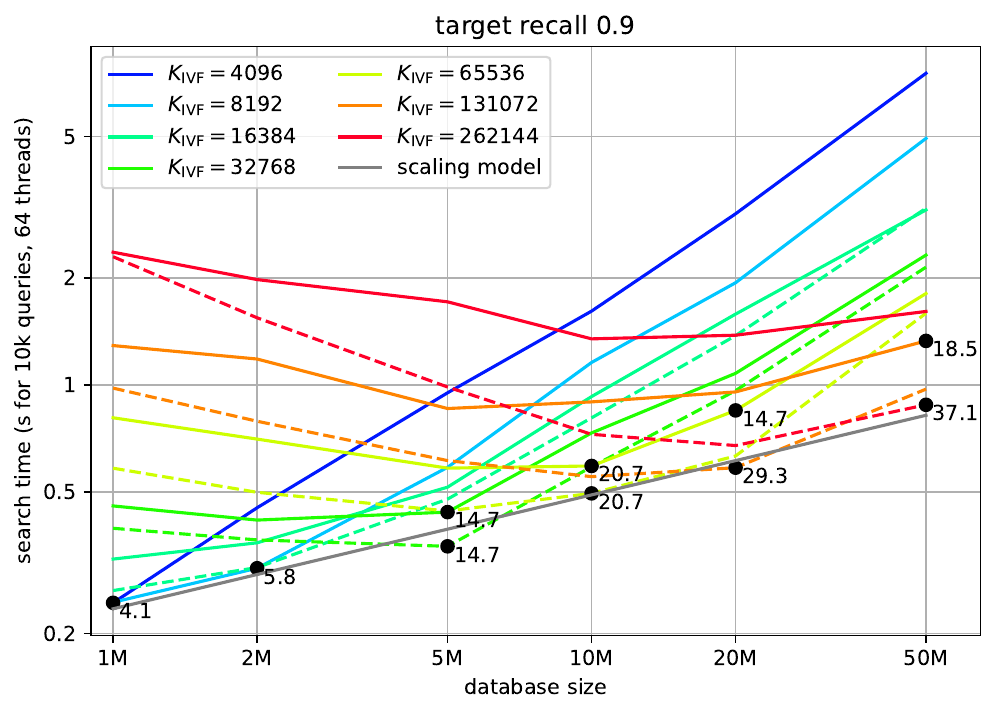}
    \caption{
    \metric{Search time} as a function of the database size $N$ for BigANN1B with different $\nlist$ settings.
    The $\nprobe$ is set so that the \metric{1-recall@1} is 90\%.
    The full lines indicate that the coarse quantizer is exact, the dashed lines rely on a HNSW coarse quantizer. 
    For some setting we indicate the ratio $\nlist / \sqrt{N}$
    }
    \label{fig:centroid_factor}
\end{figure}

Figure~\ref{fig:centroid_factor} shows the optimal settings of $\nlist$ for various database sizes. 
For a small $\nlist=4096$, the coarse quantization runtime is negligible and the \metric{search time} increases linearly with the database size. 
On larger datasets, it is beneficial to increase $\nlist$. 
As in \eqref{eq:centroidfactor}, the ratio $\nlist / \sqrt{N}$ is roughly 15 to 20. Note that this ratio depends on the data distribution and the target \metric{accuracy}.
Interestingly, in a regime where $\nlist$ is larger than the optimal setting for $N$ (\eg $\nlist=2^{18}$ and $N=$5M), the $\nprobe$ needed to reach the target accuracy \emph{decreases} with the dataset size, and so does the \metric{search time}. 
This is because when $\nlist$ is fixed and $N$ increases, for a given query vector, the nearest database vector is either the same or a new one that is closer, so it is more likely to be found in a quantization cluster nearer to the query. 

    With a faster non-exhaustive coarse quantizer (e.g. HNSW) it is even more useful to increase $\nlist$ for larger databases, as the coarse quantization becomes relatively cheap. 
At the limit, when $\nlist=N$, then all the work is done by the coarse quantizer. 
In this scenario, the limiting factor becomes the \metric{memory overhead} of the coarse quantizer.

By fitting a model of the form $ t = t_0 N^\alpha $ to the timings of the fastest index in Figure~\ref{fig:centroid_factor}, we can derive a scaling rule for the IVF indexes: 
\begin{center}
    \begin{tabular}{l|rrrr}
    \toprule
    target recall@1 & 0.5 & 0.75 & 0.9 & 0.99 \\
    power $\alpha$  &  0.29 & 0.30 & 0.34 & 0.45 \\ 
    \bottomrule
    \end{tabular}
\end{center}
Thus, with this model, the search time increases faster for higher accuracy targets, but $\alpha<0.5$, so the runtime dependence on the database size is below $\sqrt{N}$.
This empirical complexity analysis has been used before for graph-based indices~\cite{wang2021comprehensive}.

\myparagraph{Encoding residuals}
\label{sec:residuals}
In general, it is more \metric{accurate} to compress the residuals of the database vectors w.r.t. the centroids~\cite[Eq. (28)]{jegou2010product}. 
This is either because the norm of the residuals is lower than that of the original vectors, or because residual encoding is a way to take into account a-priori information from the coarse quantizer.  
In Faiss, this is controlled via the \faissfunction{IndexIVF.by\_residual} flag, which is set to true by default. 

\begin{figure}
    \centering
    \includegraphics[width=\columnwidth]{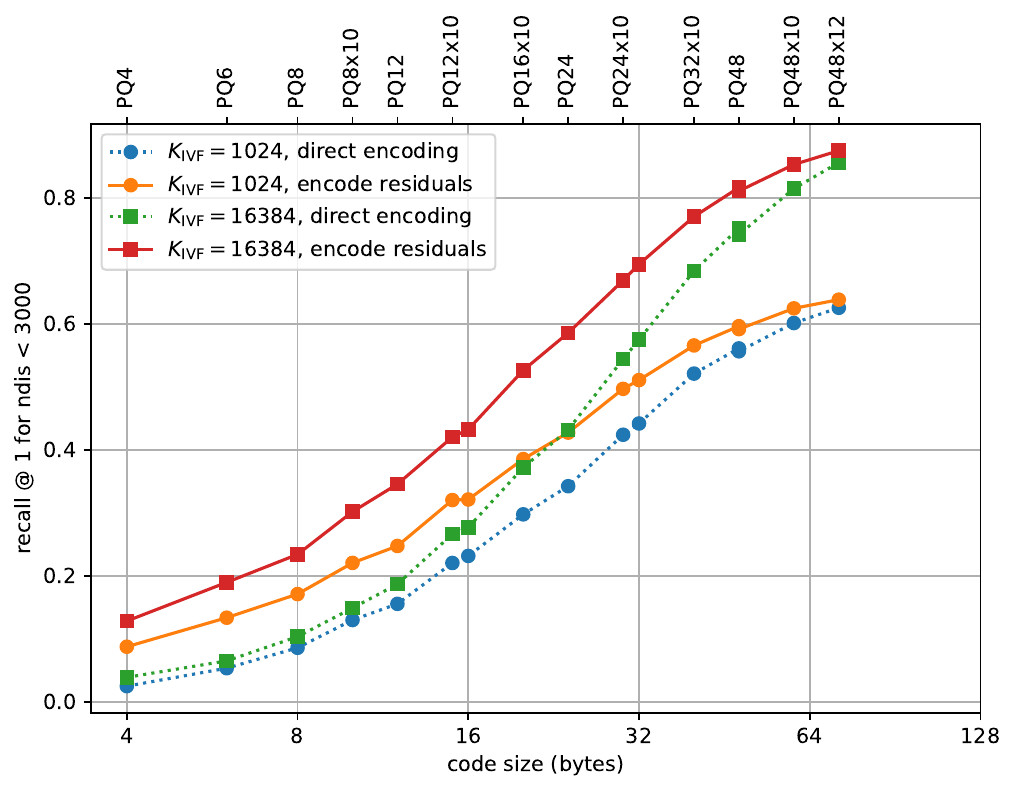}
    \caption{
    Comparing IVF indexes with and without residual encoding for $\nlist \in \{2^{10}, 2^{14}\}$ on the Deep1M dataset ($d$=96 dimensions), with different product quantization settings. 
    We measure the recall that can be achieved within 3000 distance comparisons. 
    }
    \label{fig:residuals}
\end{figure}

Figure~\ref{fig:residuals} shows that encoding residuals is beneficial for shorter codes. 
For larger codes, the contribution of the residual is less important. 
Indeed, as the original data is 96-dimensional, it can be compressed to 64 bytes relatively accurately. 
Note that using higher $\nlist$ also improves the accuracy of the quantizer with residual encoding. 
From a pure encoding point of view, the  additional bits of information brought by the coarse quantizer ($\log_2(\nlist)=10$ or $14$) improve the accuracy more when used in this residual encoding than if they would added to increase the size of a PQ.

\myparagraph{Spherical clustering for inner product search}
Efficient indexing for maximum inner product search (MIPS) faces multiple issues: 
the distribution of query vectors is often different from the database vector distribution, most notably in recommendation systems~\cite{paterek2007improving};
the MIPS datasets are diverse, an algorithm that obtains a good performance on some dataset will perform badly on another. 
Besides, \cite{morozov2018non} show that using the preprocessing formulas in Section~\ref{sec:metricequiv} is a suboptimal way of indexing for MIPS.

Several specialized clustering and indexing methods were developed for MIPS~\cite{guo2020accelerating,morozov2018non}.
Instead, Faiss implements a modification of k-means clustering, spherical k-means~\cite{dhillon2001concept}, which normalizes the IVF centroids at each iteration.
One of the MIPS issues is due to database vectors of very different norms (when they are normalized, MIPS is equivalent to L2 search). 
High-norm centroids ``attract'' the database vectors in their clusters, which increases the imbalance factor.
Spherical k-means is designed to avoid this issue. 

\begin{figure}
    \centering
    \includegraphics[width=\columnwidth]{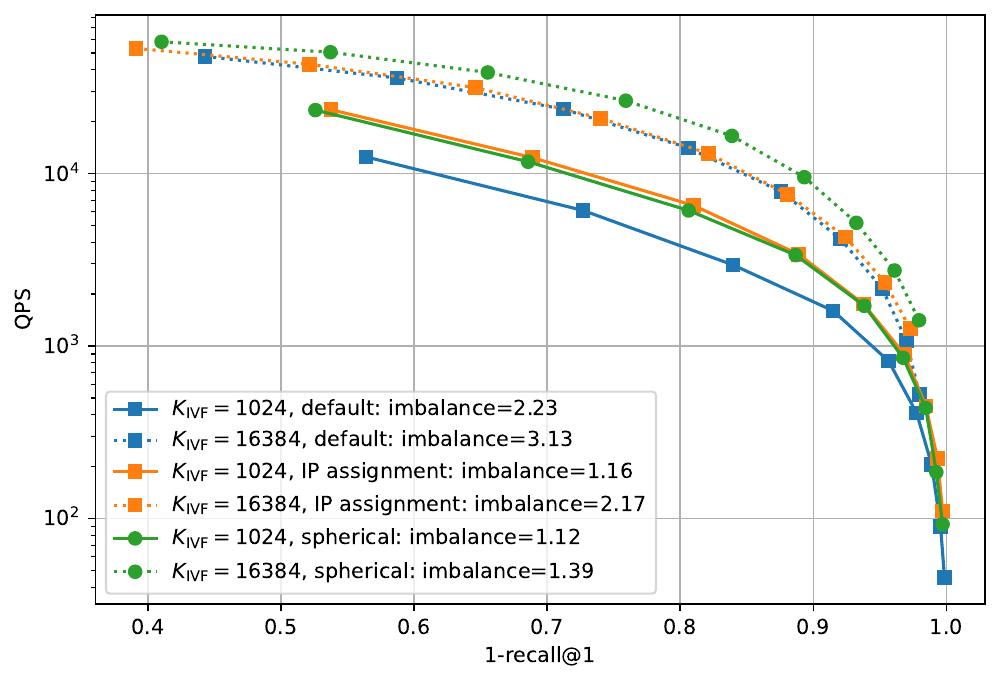}
    \caption{
    Precision vs. speed trade-off for the MIPS contriever1M dataset.
    The compared settings are whether the coarse quantizer assignement is done using L2 distance (default) or IP assignment and whether the k-means clustering does a normalization at each iteration (spherical, IP and L2 assignment are equivalent in that case). 
    The imbalance factors are indicated for each setting. 
    }
    \label{fig:mipssearch}
\end{figure}

Figure~\ref{fig:mipssearch} shows that for the Contriever MIPS dataset, the imbalance factor is high. 
It is reduced by using IP assignment instead of L2, and even more with spherical k-means.

\myparagraph{Big batch search} 
A common use case for ANNS is search with very large query batches. 
This appears for applications such as large-scale data deduplication. 
In this case, rather than loading an entire index in memory and processing queries one small batch at a time, it can be more memory-efficient to load only the quantizer, quantize the queries, and then iterate over the index by loading it one chunk at a time.
Big-batch search is implemented in the module \faissfunction{contrib.big\_batch\_search}.

\subsection{Graph based} \label{graphbased}

Graph-based indexing consists in building a directed graph whose nodes are the vectors to index. 
At search time, the graph is explored by following the edges towards the nodes that are closest to the query vector. 
In practice, the search is not greedy but maintains a priority queue with the most promising edges to explore. 
Thus, the trade-off at search time is given by the number of exploration steps: higher is more \metric{accurate} but \metric{slower}. 

A graph-based algorithm is a general framework that can encompass many variants. 
In particular, tree-based search or IVF can be seen as special cases of graphs. 
One can see graphs as a way to precompute neighbors for the database vectors, then match the query to one of the vertices and follow the neighbors from there.
However, they can also be built to handle out-of-distribution queries~\cite{jaiswal2022ooddiskann,chen2024roargraph}.

Given this search algorithm, relying on a pure k-nearest neighbor graph is not optimal because neighbors are redundant. 
Therefore, the graph building heuristic consists in balancing edges to nearest neighbors and edges that reach more distant nodes. 
Most graph methods fix the number of outgoing edges per node,  
which adjusts the trade-off between \metric{search speed} and \metric{memory usage}. 
The memory usage per vector breaks down into (1) the possibly compressed vector and (2) the outgoing edges for that vector~\cite{douze2018link}.

Faiss implements two graph-based algorithms: HNSW and NSG, respectively in the \faissfunction{IndexHNSW} and \faissfunction{IndexNSG} classes. 
 
\myparagraph{HNSW} 
The hierarchical navigable small world graph~\cite{malkov2018efficient} is a search structure where some randomly selected vertices are promoted to be hubs that are explored first. 
A notable advantage of HNSW is its ability to add vectors on-the-fly. 

\myparagraph{NSG} 
The Navigating Spreading-out Graph~\cite{fu2017fast} is built from a k-nearest neighbor graph that must be provided on input. 
At building time, some short-range edges are replaced with longer-range edges. 
The input k-nn graph can be built with a brute force algorithm or with a specialized method such as NN-descent~\cite{dong2011efficient} (\faissfunction{NNDescent}). 
Unlike HNSW, NSG does not rely on multi-layer graph structures, but uses long connections to achieve fast navigation. 
In addition, NSG starts from a fixed center point when searching.

\begin{figure}
    \centering
    \includegraphics[width=\columnwidth]{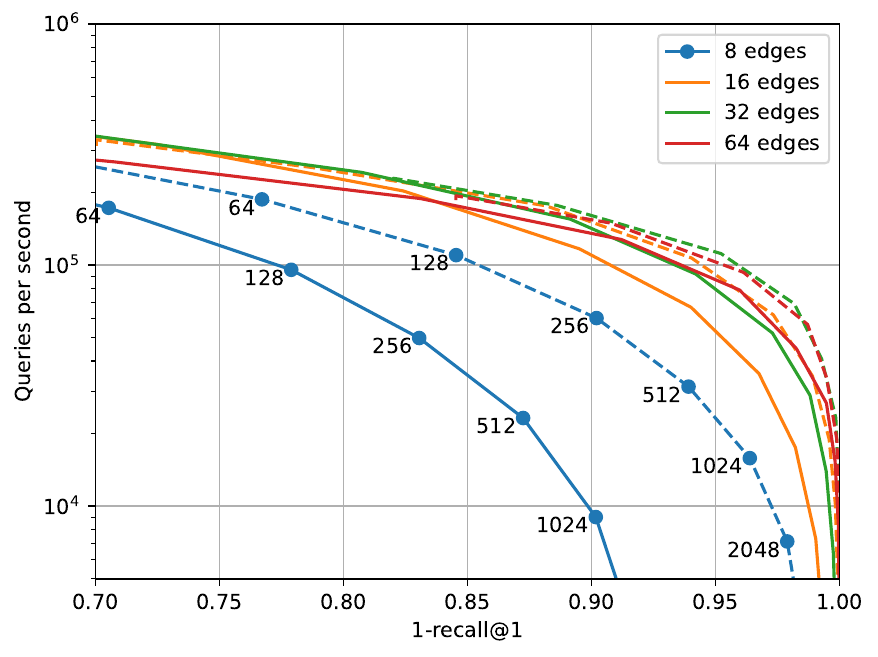}
    \caption{Comparison of graph-based indexing methods HNSW (full lines) and NSG (dashes) to index Deep1M. 
    We sweep the trade-offs between \metric{speed} and \metric{accuracy} by varying the  number of graph traversal steps (indicated for some of the curves). 
    }
    \label{fig:graph_based}
\end{figure}

\myparagraph{Discussion} 
Figure~\ref{fig:graph_based} compares the speed-accuracy trade-off for the NSG and HNSW indexes. 
Their main build-time hyperparameter is the number of edges per node, so we tested several settings (for HNSW this is the number of edges on the base level of the hierarchical graph). 
The main search-time parameter is the number of graph traversal steps during search (parameter \faissfunction{efSearch} for HNSW and \faissfunction{search\_L} for NSG), which we vary to plot each curve.
Increasing the number of edges improves the results until 64 edges, beyond which performance deteriorates. 
NSG obtains better trade-offs in general, at the cost of a longer \metric{build time}. 
Building the k-NN graph with NN-descent for 1M vectors takes 37\;s, and about the same time with exact, brute force search on a GPU.
The NSG graph is frozen after the first batch of vectors is added, there is no easy way to add more vectors afterwards.

\subsection{How to choose an index}
\label{sec:howtochoseindex}

\begin{figure}
    \centering
    \includegraphics[width=\linewidth]{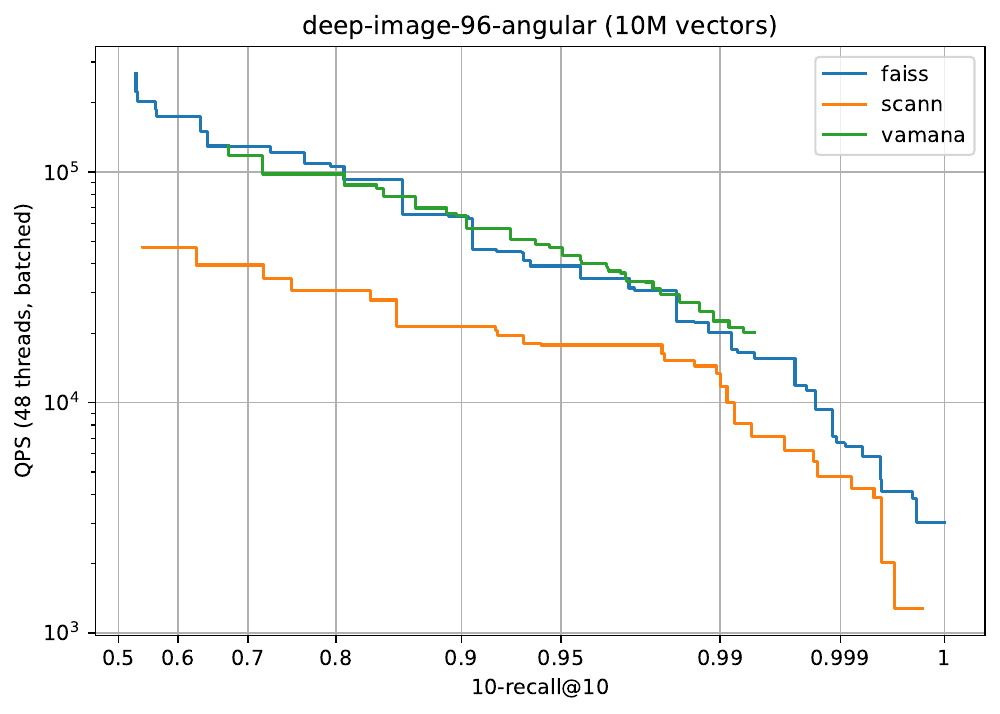}
    \caption{
    Comparison between the Faiss two other vector search libraries (SCANN and Diskann/vamana) when Deep10M in the ann-benchmarks setup (batch mode). 
    }
    \label{fig:annbenchmark}
\end{figure}

In most cases, choosing an appropriate index can be done by following a process delineated in Appendix~\ref{app:decisiontree}. 
First, one has to decide whether indexing is needed at all: indeed, in some cases a direct brute force search is the best option. 
Otherwise, the choice is between IVF and graph-based indexes. 

\myparagraph{IVF vs. graph-based} 
Graph-based indices are a good option for indexes where there is no constraint on \metric{memory usage}, typically for indexes below 1M vectors. 
Beyond 10M vectors, the \metric{construction time} typically becomes the limiting factor. 
For larger indexes, where compression is required to even fit the database vectors in memory, IVF indexes are the only option. 

The decision tree of Figure~\ref{fig:decisiontree} provides intial directions.
The Faiss wiki\footnote{\url{https://github.com/facebookresearch/faiss/wiki/Indexing-1G-vectors}} features comprehensive benchmarks for various database sizes and memory budgets.
To refine the index parameters, benchmarking should be used.

\subsection{Benchmarking indexes}
\label{sec:benchfw}

Faiss includes a benchmarking framework (\faissfunction{bench\_fw}) that optimizes index types and parameters to explore \metric{accuracy}, \metric{memory usage} and \metric{search time} operating points. 
The benchmark generates candidate index configurations to evaluate, sweeps both construction-time and search-time parameters, and measures these metrics. 

\myparagraph{Decoupling encoding and non-exhaustive search options}
Beyond a certain scale, \metric{search time} is determined by the number of distance computations performed between the query vector and database vectors.

As shown in Sections~\ref{sec:pruning} and \ref{sec:compression}, Faiss indexes are built as a combination of pruning and compression, see Table~\ref{tab:combinations}.
To evaluate index configurations efficiently, the benchmarking framework takes advantage of this compositional design. 
The training of vector transformations and k-means clustering for IVF coarse quantizers are factored out and reused when possible.
Coarse quantizers and IVF indices are first trained and evaluated separately, the parameter space is pruned as described in the previous section, and only the combinations of Pareto-optimal components are benchmarked together.

\subsection{Comparison with other libraries}

ANN-benchmarks~\cite{aumuller2020ann} is a codebase that compares several ANNS implementations. 
We use this setup to compare Faiss with SCANN~\cite{guo2020accelerating} and Vamana~\cite{subramanya2019diskann}, two other industry-standard packages for vector search.
We run the search in the following setting: batch search on the Deep10M dataset with 10-recall@10 as the metric, where the training is performed on the database vectors. This differs slightly from the evaluation protocol used originally for this dataset. 

Figure~\ref{fig:annbenchmark} shows that Faiss is faster than SCANN and about on-par with Vamana, depending on the operating point. 
For Faiss we used IVF with a Product Residual Quantizer optimized for SIMD (see Appendix~\ref{app:FastScan}), followed by re-ranking (Section~\ref{sec:refine}).

\section{Database operations}
\label{sec:vectordb}

In the experiments above, the indexes are built in one go with all the vectors, while search operations are performed with one batch containing all query vectors. 
In real settings, the index evolves over time, vectors may be dynamically added or removed, searches may have to take into account metadata, etc.
In this section we show how Faiss supports some of these operations. 
Specific APIs are available to interface with external storage (Appendix~\ref{app:externalstorage}) if fine-grained control is required.

\subsection{Identifier-based operations}

Faiss indexes support two types of identifiers: sequential and arbitrary ids. Sequential ids are based on the order of additions in the index. Alternatively, the user can provide arbitrary 63-bit integer ids associated to each vector (the sign bit is reserved for invalid results). 
The corresponding addition methods for the index are \faissfunction{add} and \faissfunction{add\_with\_ids}.
Unlike e.g. Usearch~\cite{Vardanian_USearch_2022}, Faiss does not store arbitrary metadata with the vectors.

\myparagraph{Index updates}
The Faiss API includes methods to remove vectors (\faissfunction{remove\_ids}) and update them (\faissfunction{update\_vectors}) by passing the corresponding ids. 

Note that if the vector \emph{distribution} changes significantly because of additions/removals or updates, then the efficiency of any technique that fits the data distribution degrades. 
This includes IVF and PQ compression. 
This degradation can be addressed by explicit updates of the index structure~\cite{baranchuk2023dedrift}.

\myparagraph{Flat indexes}
Sequential indexes (\faissfunction{IndexFlatCodes}) store vectors as an array. 
They support only sequential ids.
When arbitrary ids are needed, the index can be embedded in a \faissfunction{IndexIDMap}, that translates sequence numbers to arbitrary ids using a \verb|int64| array. 
This enables \faissfunction{add\_with\_ids} and returns the arbitrary ids at search time.
\faissfunction{IndexIDMap2} in addition maps arbitrary ids back to sequential ids using a hash table. 
The graph indexes rely on an embedded \faissfunction{IndexFlatCodes} to store the actual vectors. 
To use them with non-sequential ids, they should be wrapped with the same mapping objects.

\myparagraph{IVF indexes}
The IVF indexing structure supports user-provided ids natively at addition and search time.
However, id-based access may require a sequential scan, as the entries are stored in an arbitrary order in the inverted lists. 
Therefore, the IVF index can optionally maintain a \faissfunction{DirectMap}, which maps user-visible ids to the inverted list and the offset they are stored in. 
It supports lookup, removal and update by ids.
The map can be an array, which is appropriate for sequential ids, or a hash table, for arbitrary 63-bit ids. 
The direct map incurs a \metric{memory overhead} and an \metric{add-time computation overhead}, therefore, it is disabled by default. 

\myparagraph{Graph indexes}
The graph index HNSW does not support suppression and mutation, and NSG does not even support adding vectors incrementally. 
Supporting this requires heuristics to rebuild the graph when it is mutated, which are implemented in HNSWlib and FreshDiskANN \cite{singh2021freshdiskann} but suboptimal indexing-wise.

\subsection{Filtered search}

Vector filtering consists in returning only database vectors based on some search-time criterion, other vectors are ignored.  
Faiss has basic support for vector filtering: 
the user can provide a predicate (\faissfunction{IDSelector} callback), and if the predicate returns false on the vector id, the vector is ignored.

Therefore, if metadata is needed to filter the vectors, the callback function needs to do an indirection to the metadata table, which is inefficient. 
Another approach is to exploit the unused bits of the identifier. 
If $N$ documents are indexed with sequential ids, $63 - \lceil \log_2(N) \rceil$ bits are unused. 

This is sufficient to store enumerated types (\eg country codes, music genres, license types, etc.), dates (as days since some origin),  version numbers, etc. 
However, it is insufficient for more complex metadata. 
In the example use case below, we use the available bits to implement more complex filtering.

\myparagraph{Filtering with bag-of-word vectors}
In the filtered search track of the BigANN 2023 competition~\cite{bigann23}, each query and database vector is associated with a few terms from a fixed vocabulary of size $v$ (for the queries there are only 1 or 2 words).  
The filtering consists in considering only the database vectors that include all the query terms. %
. 
This metadata is given as a sparse matrix $M_\mathrm{meta} \in \{0,1\}^{N\times v}$.

The basic implementation of the filter starts from query vector $q$ and the associated words $w_1, w_2 \in \{1...v\}$. 
Before computing a distance to a vector with id $i$, it fetches row $i$ of $M_\mathrm{meta}$ to verify that $w_1$ and $w_2$ are in it.
This predicate is slow because (1) it requires to access $M_\mathrm{meta}$, which causes cache misses and (2) it performs an iterative binary search in the sparse matrix structure. 
Since the callback is called in the tightest inner loop of the search function, and since the IVF search tends to perform many vector comparisons, this has non negligible performance impact.

To speed up the predicate, we can use bit manipulations. 
In this example, $N=10^7$, so we use only $\lceil \log_2 N \rceil = 24$ bits of the ids, leaving $63-24 = 39$ bits that are always 0. 
We associate to each word $j$ a 39-bit signature $S[j]$, and to each set of words the binary ``or'' of these signatures. 
The query is represented by $s_\mathrm{q} = S[w_1] \vee S[w_2]$. 
Database entry $i$ with words $W_i$ is represented by $s_i = \vee_{w\in W_i} S[w]$. 
Then the following implication holds: if $\{w_1, w_2\} \subset W_i$ then all 1 bits of $s_\mathrm{q}$ are also set to 1 in $s_i$:
\begin{equation}
\{w_1, w_2\} \subset W_i \Rightarrow \neg s_i \wedge s_\mathrm{q} = 0,     
\end{equation}
which is equivalent to:
\begin{equation}
    \neg s_i \wedge s_\mathrm{q} \neq 0 \Rightarrow \{w_1, w_2\} \not\subset W_i. 
\end{equation}
This binary test costs only a few machine instructions on data that is already in machine registers. 
It can thus be used as a pre-filter before applying the predicate computation. 
This is implemented in the module \faissfunction{bow\_id\_selector}\footnote{\url{https://github.com/harsha-simhadri/big-ann-benchmarks/tree/main/neurips23/filter/faiss}}. 

The remaining degree of freedom is how to choose the binary signatures, because this rule's filtering ability depends on the choice of the signatures $S$. 
We experimented with i.i.d. Bernoulli bits with varying $p$: the best setting avoids running the full predicate more than 4/5 times. 
\begin{center}
\scalebox{0.85}{
    \begin{tabular}{l|cccc}
    \toprule
    $p$\,= probability of 1  & 0.05 & 0.1 & 0.2 & 0.5 \\
    Filter hit rate  & 75.4\% & 82.1\% & 76.4\% & 42.0\%  \\ 
    \bottomrule
    \end{tabular}
}
\end{center}

\myparagraph{Vector-first or metadata-first search}
There are two possible approaches to filtered search: \emph{vector-first}, which is described above, and \emph{metadata-first}, where only vectors with appropriate metadata are considered in vector search. 
The metadata-first filtering generates a subset of vectors to compare with that can then be compared using brute force search. 
Brute force search is slow but acceptable if the subset size is small, and the results are exact. 

Therefore, in the context of the BigANN competition~\cite{bigann23}, the decision to use vector-first or metadata-first depends on how large the subset is. 
To this end, we map each word $w$ to the list of items that contain $w$, of size $L_w$.

If there is a single query word $\{w_1\}$ then the subset size is directly accessible as $S = L_{w_1}$. 
With two query words $\{w_1, w_2\}$, finding the subset size requires intersecting the inverted lists for $w_1$ and $w_2$, which is slow. 
Instead, one can estimate the size of the subset using the empirical probability of each word to appear $P(w) = L_w / N$.
Assuming independent draws, the probability of both words to appear is $P(w_1)P(w_2)$. 
Therefore, the expected size of the intersection is $ S \approx NP(w_1)P(w_2) = L_{w_1}L_{w_2} / N$.
If $S/N < 3\times 10^{-4}$ (an empirical threshold), then the subset is sufficiently small that metadata-first can be applied. 
Of course, if metadata-first is selected, the actual intersection has to be computed. 
Most participants to the competition used similar heuristics.

\section{Faiss applications} 
\label{sec:applications}

Faiss is widely used across the industry, with numerous applications leveraging its capabilities. The following examples highlight notable use cases that demonstrate exceptional scalability or significant impact.

\subsection{Trillion scale index}

In this example, we index 1.5 trillion vectors in 144 dimensions.
The indexing needs to be accurate, therefore the compression of the vectors is limited to 54 bytes with a PCA to 72 dimensions and 6-bit scalar quantizer (\verb|PCAR72,SQ6|).

A HNSW coarse quantizer with 10M centroids is used for the \faissfunction{IndexIVFScalarQuantizer}, trained with a simple distributed GPU k-means (implemented in \faissfunction{faiss.clustering}). 

Once the training is completed, the index is built in the following three phases: 
\begin{enumerate}
    \item shard over ids: 
    add the input vectors in 2000 shards independently, producing 2000 indexes (each one fits in 256\;GB RAM);
    \item shard over lists: 
    build the 100 indexes corresponding each to a subset of 100k inverted lists. 
    This is done on 100 different machines, each reading from the 2000 sharded indices, and writing the results directly to a distributed file system; 
    \item load the shards: 
    memory-map all 100 indexes on a central machine as 100  \faissfunction{OnDiskInvertedLists} (a memory map of 83\;TiB). 
\end{enumerate}

Steps 1 and 2 are organized to be performed as independent cluster jobs on a few hundred servers (64 cores, 256G RAM). 
Otherwise, the code is written in standard Faiss in Python. 

The central machine that handles searches performs the coarse quantization and loads the inverted lists from the distributed disk partition. 
The limiting factor is the network bandwidth of this central machine. 
Therefore, it is more efficient to distribute the search on 20 intermediate servers to spread the load. 
This brings the search time down to roughly 1\,s per query.

\subsection{Text retrieval}

Faiss is commonly used for knowledge intensive natural language processing tasks. In particular, ANNS is relevant for information retrieval \cite{thakur2021beir,petroni-etal-2021-kilt}, with applications such as fact checking, entity linking, slot filling or open-domain question answering: these often rely on retrieving relevant content across a large-scale corpus. To that end, embedding models have been optimized for text retrieval \cite{izacard2021contriever,lin2023train}.

Finally, \cite{JMLR:v24:23-0037}, \cite{Lin2023RADITRD}, \cite{shi2023replug} and \cite{khandelwal2020generalization} consist of language models that have been trained to integrate textual retrieval in order to improve their accuracy, factuality or compute efficiency.

\subsection{Data mining}

Another recurrent application of ANNS and Faiss is in the mining and curation of large datasets. In particular, Faiss has been used to mine bilingual texts across very large text datasets retrieved from the web \cite{schwenk-douze-2017-learning, barrault2023seamlessm4t}, or to organize a language model's training corpus in order to group together series of documents covering similar topics \cite{Shi2023InContextPL}.

In the image domain, \cite{oquab2023dinov2} leverages Faiss to remove duplicates from a dataset containing 1.3B images. It then relies on efficient indexing in order to mine a curated dataset whose distribution matches the distribution of a target dataset.

\subsection{Content Moderation}

One of the major applications of Faiss is the detection and remediation of harmful content at scale. 
Human-labeled examples of policies violating images and videos are embedded with models such as SSCD~\cite{pizzi2022self} and stored in a Faiss index. 
To decide if a new image or video would violate some policies, a multi-stage classification pipeline first embeds the content and searches the Faiss index for similar labeled examples, typically utilizing range queries. 
The results are aggregated and processed through additional machine classification or human verification. 
Since the impact of mistakes is high, good representations should discriminate perceptually similar and different content, and accurate similarity search is required even at billion to trillion scale. 
The former problem motivated the Image and Video Similarity Challenges~\cite{douze20212021,pizzi2023VSC}.

\section{Conclusion}

Throughout the years, Faiss continuously expanded its focus to include the most relevant vector indexing techniques from research. 
We continue doing this to include
novel quantization techniques~\cite{huijben2024QINco},
better hardware support for some indexes~\cite{ootomo2023cagra}
and new indexing forms, such as associative vector memories for transformer architectures~\cite{chen2024magicpig,zhang2024memorymosaics}.

\bibliographystyle{plain}
\bibliography{bibliography}

\begin{thebibliography}{10}

\bibitem{Amara2022NearestPerspective}
Kenza Amara, Matthijs Douze, Alexandre Sablayrolles, and Hervé J{\'{e}}gou.
\newblock Nearest neighbor search with compact codes: A decoder perspective.
\newblock In {\em ICMR}, 2022.

\bibitem{andoni2015optimal}
Alexandr Andoni and Ilya Razenshteyn.
\newblock Optimal data-dependent hashing for approximate near neighbors.
\newblock In {\em Proceedings of the forty-seventh annual ACM symposium on
  Theory of computing}, pages 793--801, 2015.

\bibitem{aumuller2020ann}
Martin Aum{\"u}ller, Erik Bernhardsson, and Alexander Faithfull.
\newblock Ann-benchmarks: A benchmarking tool for approximate nearest neighbor
  algorithms.
\newblock {\em Information Systems}, 87:101374, 2020.

\bibitem{babenko2014additive}
Artem Babenko and Victor Lempitsky.
\newblock Additive quantization for extreme vector compression.
\newblock In {\em Conference on Computer Vision and Pattern Recognition}, 2014.

\bibitem{babenko2015tree}
Artem Babenko and Victor Lempitsky.
\newblock Tree quantization for large-scale similarity search and
  classification.
\newblock In {\em Proceedings of the IEEE Conference on Computer Vision and
  Pattern Recognition}, pages 4240--4248, 2015.

\bibitem{babenko2016efficient}
Artem Babenko and Victor Lempitsky.
\newblock Efficient indexing of billion-scale datasets of deep descriptors.
\newblock In {\em Conference on Computer Vision and Pattern Recognition}, 2016.

\bibitem{bachrach2014speeding}
Yoram Bachrach, Yehuda Finkelstein, Ran Gilad-Bachrach, Liran Katzir, Noam
  Koenigstein, Nir Nice, and Ulrich Paquet.
\newblock Speeding up the xbox recommender system using a euclidean
  transformation for inner-product spaces.
\newblock In {\em Proceedings of the 8th ACM Conference on Recommender
  systems}, pages 257--264, 2014.

\bibitem{baranchuk2023dedrift}
Dmitry Baranchuk, Matthijs Douze, Yash Upadhyay, and I~Zeki Yalniz.
\newblock Dedrift: Robust similarity search under content drift.
\newblock In {\em Proceedings of the IEEE/CVF International Conference on
  Computer Vision}, pages 11026--11035, 2023.

\bibitem{barrault2023seamlessm4t}
Lo{\"\i}c Barrault, Yu-An Chung, Mariano~Cora Meglioli, David Dale, Ning Dong,
  Paul-Ambroise Duquenne, Hady Elsahar, Hongyu Gong, Kevin Heffernan, John
  Hoffman, et~al.
\newblock Seamlessm4t-massively multilingual \& multimodal machine translation.
\newblock {\em arXiv preprint arXiv:2308.11596}, 2023.

\bibitem{bojanowski2017enriching}
Piotr Bojanowski, Edouard Grave, Armand Joulin, and Tomas Mikolov.
\newblock Enriching word vectors with subword information.
\newblock {\em Transactions of the association for computational linguistics},
  5:135--146, 2017.

\bibitem{boytsov2016off}
Leonid Boytsov, David Novak, Yury Malkov, and Eric Nyberg.
\newblock Off the beaten path: Let's replace term-based retrieval with k-nn
  search.
\newblock In {\em Proceedings of the 25th ACM international on conference on
  information and knowledge management}, pages 1099--1108, 2016.

\bibitem{bruch2023approximate}
Sebastian Bruch, Franco~Maria Nardini, Amir Ingber, and Edo Liberty.
\newblock An approximate algorithm for maximum inner product search over
  streaming sparse vectors.
\newblock {\em arXiv preprint arXiv:2301.10622}, 2023.

\bibitem{Cao_2017_ICCV}
Zhangjie Cao, Mingsheng Long, Jianmin Wang, and Philip~S. Yu.
\newblock Hashnet: Deep learning to hash by continuation.
\newblock In {\em Proceedings of the IEEE International Conference on Computer
  Vision (ICCV)}, Oct 2017.

\bibitem{caron2018deep}
Mathilde Caron, Piotr Bojanowski, Armand Joulin, and Matthijs Douze.
\newblock Deep clustering for unsupervised learning of visual features.
\newblock In {\em Proceedings of the European conference on computer vision
  (ECCV)}, pages 132--149, 2018.

\bibitem{caron2021emerging}
Mathilde Caron, Hugo Touvron, Ishan Misra, Herv{\'e} J{\'e}gou, Julien Mairal,
  Piotr Bojanowski, and Armand Joulin.
\newblock Emerging properties in self-supervised vision transformers.
\newblock In {\em Proceedings of the IEEE/CVF international conference on
  computer vision}, pages 9650--9660, 2021.

\bibitem{charikar2002similarity}
Moses Charikar.
\newblock Similarity estimation techniques from rounding algorithms.
\newblock In {\em Proc. ACM symp. Theory of computing}, 2002.

\bibitem{chen2024roargraph}
Meng Chen, Kai Zhang, Zhenying He, Yinan Jing, and X~Sean Wang.
\newblock Roargraph: A projected bipartite graph for efficient cross-modal
  approximate nearest neighbor search.
\newblock {\em arXiv preprint arXiv:2408.08933}, 2024.

\bibitem{chen2020simclr}
Ting Chen, Simon Kornblith, Mohammad Norouzi, and Geoffrey Hinton.
\newblock A simple framework for contrastive learning of visual
  representations.
\newblock In {\em Proceedings of the 37th International Conference on Machine
  Learning}, Proceedings of Machine Learning Research, pages 1597--1607. PMLR,
  2020.

\bibitem{chen2010approximate}
Yongjian Chen, Tao Guan, and Cheng Wang.
\newblock {Approximate nearest neighbor search by residual vector
  quantization}.
\newblock {\em Sensors}, 10(12):11259--11273, 2010.

\bibitem{chen2024magicpig}
Zhuoming Chen, Ranajoy Sadhukhan, Zihao Ye, Yang Zhou, Jianyu Zhang, Niklas
  Nolte, Yuandong Tian, Matthijs Douze, Leon Bottou, Zhihao Jia, et~al.
\newblock Magicpig: Lsh sampling for efficient llm generation.
\newblock {\em arXiv preprint arXiv:2410.16179}, 2024.

\bibitem{chern2022tpu}
Felix Chern, Blake Hechtman, Andy Davis, Ruiqi Guo, David Majnemer, and Sanjiv
  Kumar.
\newblock Tpu-knn: K nearest neighbor search at peak flop/s.
\newblock {\em Advances in Neural Information Processing Systems},
  35:15489--15501, 2022.

\bibitem{datar2004locality}
Mayur Datar, Nicole Immorlica, Piotr Indyk, and Vahab~S Mirrokni.
\newblock Locality-sensitive hashing scheme based on p-stable distributions.
\newblock In {\em Proceedings of the twentieth annual symposium on
  Computational geometry}, pages 253--262, 2004.

\bibitem{Deng2019arcface}
Jiankang Deng, Jia Guo, Niannan Xue, and Stefanos Zafeiriou.
\newblock Arcface: Additive angular margin loss for deep face recognition.
\newblock In {\em Proceedings of the IEEE/CVF Conference on Computer Vision and
  Pattern Recognition (CVPR)}, June 2019.

\bibitem{devlin2018bert}
Jacob Devlin, Ming-Wei Chang, Kenton Lee, and Kristina Toutanova.
\newblock Bert: Pre-training of deep bidirectional transformers for language
  understanding.
\newblock {\em arXiv preprint arXiv:1810.04805}, 2018.

\bibitem{dhillon2001concept}
Inderjit~S Dhillon and Dharmendra~S Modha.
\newblock Concept decompositions for large sparse text data using clustering.
\newblock {\em Machine learning}, 42:143--175, 2001.

\bibitem{dong2011efficient}
Wei Dong, Charikar Moses, and Kai Li.
\newblock Efficient k-nearest neighbor graph construction for generic
  similarity measures.
\newblock In {\em Proceedings of the 20th international conference on World
  wide web}, pages 577--586, 2011.

\bibitem{douze2014yael}
Matthijs Douze and Herv{\'e} J{\'e}gou.
\newblock The yael library.
\newblock In {\em Proceedings of the 22nd ACM international conference on
  Multimedia}, pages 687--690, 2014.

\bibitem{douze2016polysemous}
Matthijs Douze, Herv{\'e} J{\'e}gou, and Florent Perronnin.
\newblock Polysemous codes.
\newblock In {\em European Conference on Computer Vision}, 2016.

\bibitem{douze2018link}
Matthijs Douze, Alexandre Sablayrolles, and Herv{\'e} J{\'e}gou.
\newblock Link and code: Fast indexing with graphs and compact regression
  codes.
\newblock In {\em Proceedings of the IEEE conference on computer vision and
  pattern recognition}, pages 3646--3654, 2018.

\bibitem{douze20212021}
Matthijs Douze, Giorgos Tolias, Ed~Pizzi, Zo{\"e} Papakipos, Lowik Chanussot,
  Filip Radenovic, Tomas Jenicek, Maxim Maximov, Laura Leal-Taix{\'e}, Ismail
  Elezi, et~al.
\newblock The 2021 image similarity dataset and challenge.
\newblock {\em arXiv preprint arXiv:2106.09672}, 2021.

\bibitem{duquenne2023sentence}
Paul-Ambroise Duquenne, Holger Schwenk, and Beno{\^\i}t Sagot.
\newblock Sentence-level multimodal and language-agnostic representations.
\newblock {\em arXiv preprint arXiv:2308.11466}, 2023.

\bibitem{fu2017fast}
Cong Fu, Chao Xiang, Changxu Wang, and Deng Cai.
\newblock Fast approximate nearest neighbor search with the navigating
  spreading-out graph.
\newblock {\em arXiv preprint arXiv:1707.00143}, 2017.

\bibitem{ge2013optimized}
Tiezheng Ge, Kaiming He, Qifa Ke, and Jian Sun.
\newblock Optimized product quantization for approximate nearest neighbor
  search.
\newblock In {\em Conference on Computer Vision and Pattern Recognition}, 2013.

\bibitem{gollapudi2023filtered}
Siddharth Gollapudi, Neel Karia, Varun Sivashankar, Ravishankar Krishnaswamy,
  Nikit Begwani, Swapnil Raz, Yiyong Lin, Yin Zhang, Neelam Mahapatro,
  Premkumar Srinivasan, Amit Singh, and Harsha~Vardhan Simhadri.
\newblock Filtered-diskann: Graph algorithms for approximate nearest neighbor
  search with filters.
\newblock In {\em Proceedings of the ACM Web Conference 2023}, 2023.

\bibitem{gong2012iterative}
Yunchao Gong, Svetlana Lazebnik, Albert Gordo, and Florent Perronnin.
\newblock Iterative quantization: A procrustean approach to learning binary
  codes for large-scale image retrieval.
\newblock {\em {\sc IEEE} Trans. Pattern Analysis and Machine Intelligence},
  2012.

\bibitem{guo2020accelerating}
Ruiqi Guo, Philip Sun, Erik Lindgren, Quan Geng, David Simcha, Felix Chern, and
  Sanjiv Kumar.
\newblock Accelerating large-scale inference with anisotropic vector
  quantization.
\newblock In {\em International Conference on Machine Learning}. PMLR, 2020.

\bibitem{hong2019asymmetric}
Weixiang Hong, Xueyan Tang, Jingjing Meng, and Junsong Yuan.
\newblock Asymmetric mapping quantization for nearest neighbor search.
\newblock {\em IEEE Transactions on Pattern Analysis and Machine Intelligence},
  42(7):1783--1790, 2019.

\bibitem{huang2020embedding}
Jui-Ting Huang, Ashish Sharma, Shuying Sun, Li~Xia, David Zhang, Philip Pronin,
  Janani Padmanabhan, Giuseppe Ottaviano, and Linjun Yang.
\newblock Embedding-based retrieval in facebook search.
\newblock In {\em Proceedings of the 26th ACM SIGKDD International Conference
  on Knowledge Discovery \& Data Mining}, pages 2553--2561, 2020.

\bibitem{huijben2024QINco}
Iris~A.M. Huijben, Matthijs Douze, Matthew~J. Muckley, Ruud~J.G. van Sloun, and
  Jakob Verbeek.
\newblock Residual quantization with implicit neural codebooks.
\newblock In {\em International Conference on Machine Learning (ICML)}, 2024.

\bibitem{izacard2021contriever}
Gautier Izacard, Mathilde Caron, Lucas Hosseini, Sebastian Riedel, Piotr
  Bojanowski, Armand Joulin, and Edouard Grave.
\newblock Unsupervised dense information retrieval with contrastive learning,
  2021.

\bibitem{JMLR:v24:23-0037}
Gautier Izacard, Patrick Lewis, Maria Lomeli, Lucas Hosseini, Fabio Petroni,
  Timo Schick, Jane Dwivedi-Yu, Armand Joulin, Sebastian Riedel, and Edouard
  Grave.
\newblock Atlas: Few-shot learning with retrieval augmented language models.
\newblock {\em Journal of Machine Learning Research}, 24(251):1--43, 2023.

\bibitem{jaiswal2022ooddiskann}
Shikhar Jaiswal, Ravishankar Krishnaswamy, Ankit Garg, Harsha~Vardhan Simhadri,
  and Sheshansh Agrawal.
\newblock Ood-diskann: Efficient and scalable graph anns for
  out-of-distribution queries, 2022.

\bibitem{jegou2008hamming}
Herve Jegou, Matthijs Douze, and Cordelia Schmid.
\newblock Hamming embedding and weak geometric consistency for large scale
  image search.
\newblock In {\em Computer Vision--ECCV 2008: 10th European Conference on
  Computer Vision, Marseille, France, October 12-18, 2008, Proceedings, Part I
  10}, pages 304--317. Springer, 2008.

\bibitem{jegou2010product}
Herv\'e J\'egou, Matthijs Douze, and Cordelia Schmid.
\newblock Product quantization for nearest neighbor search.
\newblock {\em {\sc IEEE} Trans. Pattern Analysis and Machine Intelligence},
  2010.

\bibitem{jegou2011aggregating}
Herv{\'e} J{\'e}gou, Florent Perronnin, Matthijs Douze, Jorge S{\'a}nchez,
  Patrick P{\'e}rez, and Cordelia Schmid.
\newblock Aggregating local image descriptors into compact codes.
\newblock {\em IEEE transactions on pattern analysis and machine intelligence},
  34(9):1704--1716, 2011.

\bibitem{jegou2011searching}
Herv{\'e} J{\'e}gou, Romain Tavenard, Matthijs Douze, and Laurent Amsaleg.
\newblock Searching in one billion vectors: re-rank with source coding.
\newblock In {\em International Conference on Acoustics, Speech, and Signal
  Processing}, 2011.

\bibitem{johnson2019billion}
Jeff Johnson, Matthijs Douze, and Herv{\'e} J{\'e}gou.
\newblock Billion-scale similarity search with {GPUs}.
\newblock {\em IEEE Trans. on Big Data}, 2019.

\bibitem{khandelwal2020generalization}
Urvashi Khandelwal, Omer Levy, Dan Jurafsky, Luke Zettlemoyer, and Mike Lewis.
\newblock Generalization through memorization: Nearest neighbor language
  models, 2020.

\bibitem{lin2023train}
Sheng-Chieh Lin, Akari Asai, Minghan Li, Barlas Oguz, Jimmy Lin, Yashar Mehdad,
  Wen tau Yih, and Xilun Chen.
\newblock How to train your dragon: Diverse augmentation towards generalizable
  dense retrieval, 2023.

\bibitem{Lin2023RADITRD}
Xi~Victoria Lin, Xilun Chen, Mingda Chen, Weijia Shi, Maria Lomeli, Rich James,
  Pedro Rodriguez, Jacob Kahn, Gergely Szilvasy, Mike Lewis, Luke Zettlemoyer,
  and Scott Yih.
\newblock Ra-dit: Retrieval-augmented dual instruction tuning.
\newblock {\em ArXiv}, abs/2310.01352, 2023.

\bibitem{liu2015improved}
Shicong Liu, Hongtao Lu, and Junru Shao.
\newblock Improved residual vector quantization for high-dimensional
  approximate nearest neighbor search.
\newblock {\em arXiv preprint arXiv:1509.05195}, 2015.

\bibitem{lloyd1982least}
Stuart Lloyd.
\newblock Least squares quantization in {PCM}.
\newblock {\em {\sc IEEE} Transactions on Information Theory}, 1982.

\bibitem{lowe2004distinctive}
David~G Lowe.
\newblock Distinctive image features from scale-invariant keypoints.
\newblock {\em International Journal of Computer Vision}, 60(2), 2004.

\bibitem{lv2004image}
Qin Lv, Moses Charikar, and Kai Li.
\newblock Image similarity search with compact data structures.
\newblock In {\em Proceedings of the thirteenth ACM international conference on
  Information and knowledge management}, pages 208--217, 2004.

\bibitem{malkov2018efficient}
Yu~A Malkov and Dmitry~A Yashunin.
\newblock Efficient and robust approximate nearest neighbor search using
  hierarchical navigable small world graphs.
\newblock {\em IEEE transactions on pattern analysis and machine intelligence},
  42(4):824--836, 2018.

\bibitem{martinez2016revisiting}
Julieta Martinez, Joris Clement, Holger~H Hoos, and James~J Little.
\newblock Revisiting additive quantization.
\newblock In {\em European Conference on Computer Vision}, 2016.

\bibitem{martinez2018lsq++}
Julieta Martinez, Shobhit Zakhmi, Holger~H Hoos, and James~J Little.
\newblock {LSQ++:} lower running time and higher recall in multi-codebook
  quantization.
\newblock In {\em European Conference on Computer Vision}, 2018.

\bibitem{matsui2018survey}
Yusuke Matsui, Yusuke Uchida, Herv{\'e} J{\'e}gou, and Shin'ichi Satoh.
\newblock A survey of product quantization.
\newblock {\em ITE Transactions on Media Technology and Applications}, 2018.

\bibitem{mikolov2013efficient}
Tomas Mikolov, Kai Chen, Greg Corrado, and Jeffrey Dean.
\newblock Efficient estimation of word representations in vector space.
\newblock {\em arXiv preprint arXiv:1301.3781}, 2013.

\bibitem{morozov2018non}
Stanislav Morozov and Artem Babenko.
\newblock Non-metric similarity graphs for maximum inner product search.
\newblock {\em Advances in Neural Information Processing Systems}, 31, 2018.

\bibitem{Morozov2019UnsupervisedSearch}
Stanislav Morozov and Artem Babenko.
\newblock Unsupervised neural quantization for compressed-domain similarity
  search.
\newblock In {\em International Conference on Computer Vision}, 8 2019.

\bibitem{muja2014scalable}
Marius Muja and David~G Lowe.
\newblock Scalable nearest neighbor algorithms for high dimensional data.
\newblock {\em IEEE transactions on pattern analysis and machine intelligence},
  36(11):2227--2240, 2014.

\bibitem{ootomo2023cagra}
Hiroyuki Ootomo, Akira Naruse, Corey Nolet, Ray Wang, Tamas Feher, and Yong
  Wang.
\newblock Cagra: Highly parallel graph construction and approximate nearest
  neighbor search for gpus, 2023.

\bibitem{oquab2023dinov2}
Maxime Oquab, Timothée Darcet, Theo Moutakanni, Huy~V. Vo, Marc Szafraniec,
  Vasil Khalidov, Pierre Fernandez, Daniel Haziza, Francisco Massa, Alaaeldin
  El-Nouby, Russell Howes, Po-Yao Huang, Hu~Xu, Vasu Sharma, Shang-Wen Li,
  Wojciech Galuba, Mike Rabbat, Mido Assran, Nicolas Ballas, Gabriel Synnaeve,
  Ishan Misra, Herve Jegou, Julien Mairal, Patrick Labatut, Armand Joulin, and
  Piotr Bojanowski.
\newblock {DINOv2}: Learning robust visual features without supervision, 2023.

\bibitem{paterek2007improving}
Arkadiusz Paterek.
\newblock Improving regularized singular value decomposition for collaborative
  filtering.
\newblock In {\em Proceedings of KDD cup and workshop}, 2007.

\bibitem{pauleve2010locality}
Lo{\"\i}c Paulev{\'e}, Herv{\'e} J{\'e}gou, and Laurent Amsaleg.
\newblock Locality sensitive hashing: A comparison of hash function types and
  querying mechanisms.
\newblock {\em Pattern recognition letters}, 31(11):1348--1358, 2010.

\bibitem{petroni-etal-2021-kilt}
Fabio Petroni, Aleksandra Piktus, Angela Fan, Patrick Lewis, Majid Yazdani,
  Nicola De~Cao, James Thorne, Yacine Jernite, Vladimir Karpukhin, Jean
  Maillard, Vassilis Plachouras, Tim Rockt{\"a}schel, and Sebastian Riedel.
\newblock {KILT}: a benchmark for knowledge intensive language tasks.
\newblock In {\em Proceedings of the 2021 Conference of the North American
  Chapter of the Association for Computational Linguistics: Human Language
  Technologies}, Online, June 2021. Association for Computational Linguistics.

\bibitem{pizzi2023VSC}
Ed~Pizzi, Giorgos Kordopatis-Zilos, Hiral Patel, Gheorghe Postelnicu,
  Sugosh~Nagavara Ravindra, Akshay Gupta, Symeon Papadopoulos, Giorgos Tolias,
  and Matthijs Douze.
\newblock The 2023 video similarity dataset and challenge.
\newblock {\em arXiv preprint arXiv:2306.09489}, 2023.

\bibitem{pizzi2022self}
Ed~Pizzi, Sreya~Dutta Roy, Sugosh~Nagavara Ravindra, Priya Goyal, and Matthijs
  Douze.
\newblock A self-supervised descriptor for image copy detection.
\newblock In {\em Proceedings of the IEEE/CVF Conference on Computer Vision and
  Pattern Recognition}, pages 14532--14542, 2022.

\bibitem{radford2021learning}
Alec Radford, Jong~Wook Kim, Chris Hallacy, Aditya Ramesh, Gabriel Goh,
  Sandhini Agarwal, Girish Sastry, Amanda Askell, Pamela Mishkin, Jack Clark,
  et~al.
\newblock Learning transferable visual models from natural language
  supervision.
\newblock In {\em International conference on machine learning}, pages
  8748--8763. PMLR, 2021.

\bibitem{sandhawalia2010searching}
Harsimrat Sandhawalia and Herv{\'e} J{\'e}gou.
\newblock Searching with expectations.
\newblock In {\em International Conference on Acoustics, Speech, and Signal
  Processing}, 2010.

\bibitem{schwenk-douze-2017-learning}
Holger Schwenk and Matthijs Douze.
\newblock Learning joint multilingual sentence representations with neural
  machine translation.
\newblock In {\em Proceedings of the 2nd Workshop on Representation Learning
  for {NLP}}, pages 157--167, Vancouver, Canada, August 2017. Association for
  Computational Linguistics.

\bibitem{Shi2023InContextPL}
Weijia Shi, Sewon Min, Maria Lomeli, Chunting Zhou, Margaret Li, Victoria Lin,
  Noah~A. Smith, Luke Zettlemoyer, Scott Yih, and Mike Lewis.
\newblock In-context pretraining: Language modeling beyond document boundaries.
\newblock {\em ArXiv}, abs/2310.10638, 2023.

\bibitem{shi2023replug}
Weijia Shi, Sewon Min, Michihiro Yasunaga, Minjoon Seo, Rich James, Mike Lewis,
  Luke Zettlemoyer, and Wen tau Yih.
\newblock Replug: Retrieval-augmented black-box language models, 2023.

\bibitem{bigann23}
Harsha~Vardhan Simhadri, Martin Aum{\"u}ller, Amir Ingber, Matthijs Douze,
  George Williams, Magdalen~Dobson Manohar, Dmitry Baranchuk, Edo Liberty,
  Frank Liu, Ben Landrum, et~al.
\newblock Results of the big ann: Neurips'23 competition.
\newblock {\em arXiv preprint arXiv:2409.17424}, 2024.

\bibitem{simhadri2022results}
Harsha~Vardhan Simhadri, George Williams, Martin Aum{\"u}ller, Matthijs Douze,
  Artem Babenko, Dmitry Baranchuk, Qi~Chen, Lucas Hosseini, Ravishankar
  Krishnaswamny, Gopal Srinivasa, et~al.
\newblock Results of the neurips’21 challenge on billion-scale approximate
  nearest neighbor search.
\newblock In {\em NeurIPS 2021 Competitions and Demonstrations Track}, pages
  177--189. PMLR, 2022.

\bibitem{pmlr-v176-simhadri22a}
Harsha~Vardhan Simhadri, George Williams, Martin Aum\"uller, Matthijs Douze,
  Artem Babenko, Dmitry Baranchuk, Qi~Chen, Lucas Hosseini, Ravishankar
  Krishnaswamny, Gopal Srinivasa, Suhas~Jayaram Subramanya, and Jingdong Wang.
\newblock Results of the neurips{’}21 challenge on billion-scale approximate
  nearest neighbor search.
\newblock In {\em Proceedings of the NeurIPS 2021 Competitions and
  Demonstrations Track}, volume 176 of {\em Proceedings of Machine Learning
  Research}, pages 177--189. PMLR, 06--14 Dec 2022.

\bibitem{singh2021freshdiskann}
Aditi Singh, Suhas~Jayaram Subramanya, Ravishankar Krishnaswamy, and
  Harsha~Vardhan Simhadri.
\newblock Freshdiskann: A fast and accurate graph-based ann index for streaming
  similarity search, 2021.

\bibitem{subramanya2019diskann}
Suhas~Jayaram Subramanya, Rohan Kadekodi, Ravishankar Krishaswamy, and
  Harsha~Vardhan Simhadri.
\newblock Diskann: Fast accurate billion-point nearest neighbor search on a
  single node.
\newblock In {\em Neurips}, 2019.

\bibitem{sun2023automating}
Philip Sun, Ruiqi Guo, and Sanjiv Kumar.
\newblock Automating nearest neighbor search configuration with constrained
  optimization.
\newblock {\em arXiv preprint arXiv:2301.01702}, 2023.

\bibitem{sun2023soar}
Philip Sun, David Simcha, Dave Dopson, Ruiqi Guo, and Sanjiv Kumar.
\newblock Soar: Improved quantization for approximate nearest neighbor search.
\newblock In {\em Thirty-seventh Conference on Neural Information Processing
  Systems}, 2023.

\bibitem{szegedy2015going}
Christian Szegedy, Wei Liu, Yangqing Jia, Pierre Sermanet, Scott Reed, Dragomir
  Anguelov, Dumitru Erhan, Vincent Vanhoucke, and Andrew Rabinovich.
\newblock Going deeper with convolutions.
\newblock In {\em Proceedings of the IEEE conference on computer vision and
  pattern recognition}, pages 1--9, 2015.

\bibitem{szilvasy2024vectorsearchsmallradiuses}
Gergely Szilvasy, Pierre-Emmanuel Mazaré, and Matthijs Douze.
\newblock Vector search with small radiuses, 2024.

\bibitem{tavenard2011balancing}
Romain Tavenard, Herv{\'e} J{\'e}gou, and Laurent Amsaleg.
\newblock Balancing clusters to reduce response time variability in large scale
  image search.
\newblock In {\em 2011 9th International Workshop on Content-Based Multimedia
  Indexing (CBMI)}, pages 19--24. IEEE, 2011.

\bibitem{thakur2021beir}
Nandan Thakur, Nils Reimers, Andreas R{\"u}ckl{\'e}, Abhishek Srivastava, and
  Iryna Gurevych.
\newblock {BEIR}: A heterogeneous benchmark for zero-shot evaluation of
  information retrieval models.
\newblock In {\em Thirty-fifth Conference on Neural Information Processing
  Systems Datasets and Benchmarks Track (Round 2)}, 2021.

\bibitem{vanluijt2020bringing}
Bob van Luijt and Micha Verhagen.
\newblock Bringing semantic knowledge graph technology to your data.
\newblock {\em IEEE Software}, 37(2):89--94, 2020.

\bibitem{Vardanian_USearch_2022}
Ash Vardanian.
\newblock {USearch by Unum Cloud}, June 2022.

\bibitem{2020SciPy-NMeth}
Pauli Virtanen, Ralf Gommers, Travis~E. Oliphant, Matt Haberland, Tyler Reddy,
  David Cournapeau, Evgeni Burovski, Pearu Peterson, Warren Weckesser, Jonathan
  Bright, St{\'e}fan~J. {van der Walt}, Matthew Brett, Joshua Wilson, K.~Jarrod
  Millman, Nikolay Mayorov, Andrew R.~J. Nelson, Eric Jones, Robert Kern, Eric
  Larson, C~J Carey, {\.I}lhan Polat, Yu~Feng, Eric~W. Moore, Jake
  {VanderPlas}, Denis Laxalde, Josef Perktold, Robert Cimrman, Ian Henriksen,
  E.~A. Quintero, Charles~R. Harris, Anne~M. Archibald, Ant{\^o}nio~H. Ribeiro,
  Fabian Pedregosa, Paul {van Mulbregt}, and {SciPy 1.0 Contributors}.
\newblock {{SciPy} 1.0: Fundamental Algorithms for Scientific Computing in
  Python}.
\newblock {\em Nature Methods}, 17:261--272, 2020.

\bibitem{wang2021milvus}
Jianguo Wang, Xiaomeng Yi, Rentong Guo, Hai Jin, Peng Xu, Shengjun Li, Xiangyu
  Wang, Xiangzhou Guo, Chengming Li, Xiaohai Xu, et~al.
\newblock Milvus: A purpose-built vector data management system.
\newblock In {\em Proceedings of the 2021 International Conference on
  Management of Data}, pages 2614--2627, 2021.

\bibitem{wang2017survey}
Jingdong Wang, Ting Zhang, Nicu Sebe, Heng~Tao Shen, et~al.
\newblock A survey on learning to hash.
\newblock {\em IEEE transactions on pattern analysis and machine intelligence},
  40(4):769--790, 2017.

\bibitem{wang2015learning}
Jun Wang, Wei Liu, Sanjiv Kumar, and Shih-Fu Chang.
\newblock Learning to hash for indexing big data - a survey.
\newblock {\em Proc. of the IEEE}, 2015.

\bibitem{wang2021comprehensive}
Mengzhao Wang, Xiaoliang Xu, Qiang Yue, and Yuxiang Wang.
\newblock A comprehensive survey and experimental comparison of graph-based
  approximate nearest neighbor search.
\newblock {\em arXiv preprint arXiv:2101.12631}, 2021.

\bibitem{weber1998quantitative}
Roger Weber, Hans-J{\"o}rg Schek, and Stephen Blott.
\newblock A quantitative analysis and performance study for similarity-search
  methods in high-dimensional spaces.
\newblock In {\em VLDB}, volume~98, pages 194--205, 1998.

\bibitem{zhang2024memorymosaics}
Jianyu Zhang, Niklas Nolte, Ranajoy Sadhukhan, Beidi Chen, and Léon Bottou.
\newblock Memory mosaics, 2024.

\end{thebibliography}

\newpage 
\appendix
\section{Appendix}

This appendix exposes aspects of Faiss's implementation. 
Faiss started in a research environment. 
As a consequence, it grew organically as indexing research was making progress.

In the following, 
we summarize the guiding principles that keep the library coherent (Appendix~\ref{app:codestruct});
the structure of the library and its dependencies (Appendix \ref{app:highlevel});
how optimization is performed (Appendix~\ref{app:optim});
an example of how Faiss internals are exposed so that it can be embedded in a vector database (Appendix~\ref{app:externalstorage}; 
and finally, a flowchart that shows how to choose a Faiss index (Appendix~\ref{app:decisiontree}).

\subsection{Code structure}
\label{app:codestruct}

The core of Faiss is implemented in C++.
The guiding principles are 
(1) the code should be as open as possible, so that users can access all the implementation details of the indexes; 
(2) Faiss should be easy to embed from external libraries;
(3) the core library focuses on vector search only.

Therefore, all fields of the classes are public (C++ \verb|struct|). 
Faiss is a late adopter for C++ standards, so that it can be used with relatively old compilers (currently C++17). 

Faiss's basic data types are concrete (not templates): vectors are always represented as 32-bit floats that are portable and provide a good trade-off between \metric{size} and \metric{accuracy}. 
Similarly, all vector ids are represented with 64-bit integers. 
This is often larger than necessary for sequential numbering but is widely used for database identifiers.

Faiss is modular and includes few dependencies, so that linking it from C++ is easy. 
Callback classes (\faissfunction{ResultHanlder}, \faissfunction{InvertedLists}, \faissfunction{IDSelector}) can be subclassed to customize the indices. 
On the Python level, they can be wrapped in SWIG to be provided to Faiss without recompiling Faiss itself.

\subsection{High-level interface}
\label{app:highlevel}

\begin{figure}
    \centering
    \includegraphics[width=\linewidth]{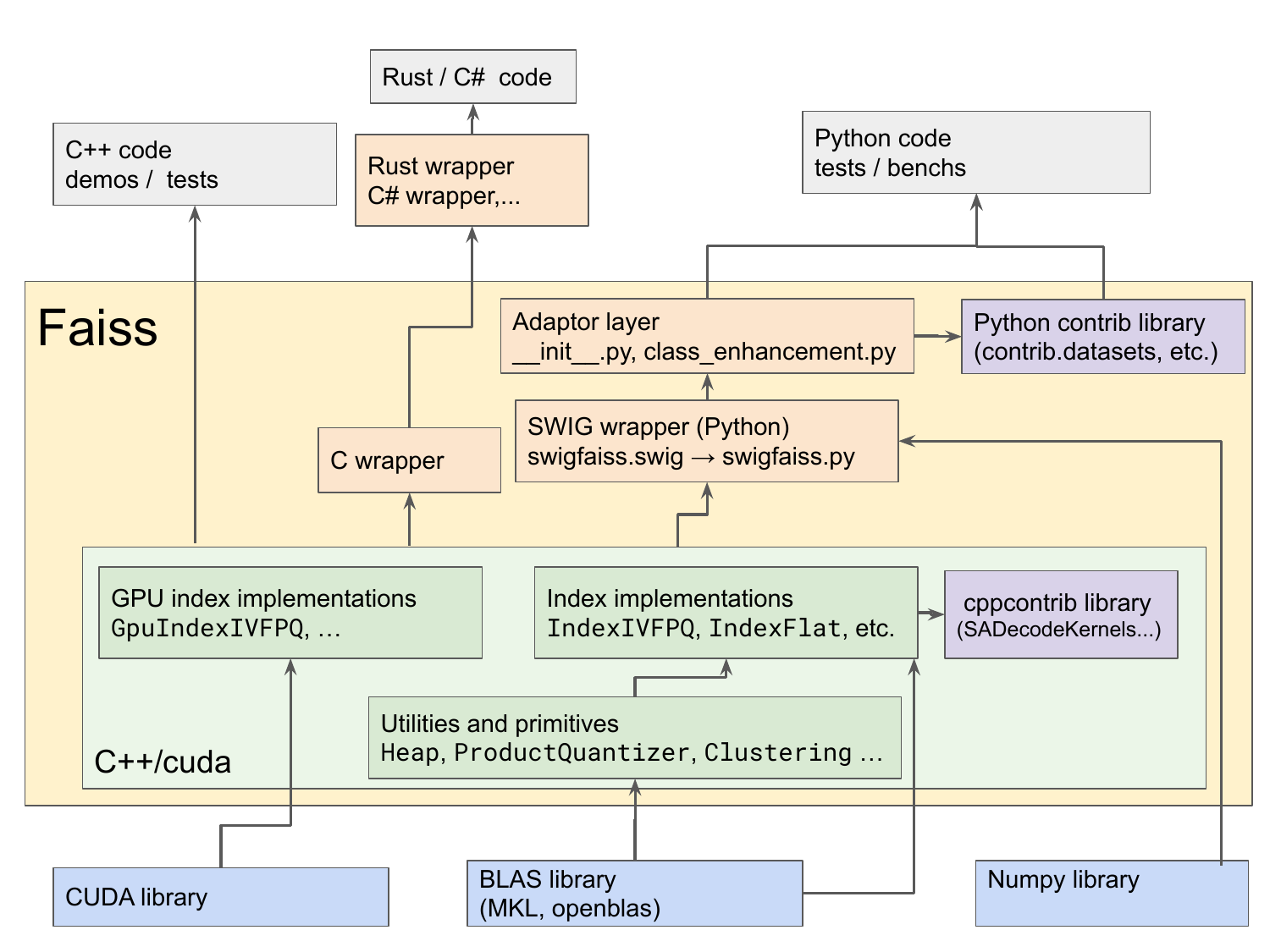}
    \caption{Architecture of the Faiss library. Arrows indicate dependencies. 
    \emph{Bottom:} the library's dependencies. \emph{Top:} example of software that depends on Faiss, most notably its extensive test suite.
    }
    \label{fig:faissdiagram}
\end{figure}

Figure~\ref{fig:faissdiagram} shows the structure of the library. 
The C++ core library and the GPU add-on have as few dependencies as possible: only a BLAS implementation and CUDA itself. 

In order to facilitate experimentation, the whole library is wrapped for Python with numpy. 
To this end, SWIG\footnote{https://www.swig.org/} exhaustively generates wrappers for all C++ classes, methods and variables.
The associated Python layer also contains benchmarking code, dataset definitions, driver code. 
More and more functionality is embedded in the \faissfunction{contrib} package of Faiss. 
Faiss also provides a pure C API, which is useful for bindings with programming languages such as Rust or Java.

The \faissfunction{Index} is presented to the end user as a monolithic object, even when it embeds other indexes as quantizers, refinement indexes or sharded sub-indexes.
Therefore, an index can be duplicated with \faissfunction{clone\_index} and serialized into as a single byte stream using a single function, \faissfunction{write\_index}. 
It also contains the necessary headers so that it can be read by a generic function, \faissfunction{read\_index}.

\myparagraph{The index factory}
Index objects can be instantiated explicitly  in C++ or Python, but it is more common to build them with the \faissfunction{index\_factory} function. 
This function takes a string that describes the index structure and its main parameters. 
For example, the string \verb|PCA160,IVF20000_HNSW,| \verb|PQ20x10,RFlat| instantiates an IVF index with $\nlist=20000$, where the coarse quantizer is a HNSW index; then the vectors are represented with a \verb|PQ20x10| product quantizer. %
The data is preprocessed with a PCA to 160 dimensions, and the search results are re-ranked with a refinement index that performs exact distance computations.
All the index parameters are set to reasonable defaults, \eg the PQ encodes the residual of the vectors w.r.t. the coarse quantization centroids.
Faiss indexes can be used as vector codecs with functions \faissfunction{sa\_encode}, \faissfunction{sa\_decode} and \faissfunction{sa\_code\_size}. %

\subsection{Optimization}
\label{app:optim}

\myparagraph{Approach to optimization}
Faiss aims at being feature complete first. 
A non-optimal version of all indexes is implemented first. 
Code is optimized only when it appears that \metric{runtime} is important for a certain index. 
The non-optimized setting is used to control the correctness of the optimized version. 

Often, only a subset of data sizes are optimized. 
For example, for PQ indexes, only $K$\,=\,$2^8$ and $K$\,=\,$2^4$ and $d/M \in \{2,4,8,16,20\}$ are fully optimized. 
For \faissfunction{IndexLSH} search, we only optimized code sizes 4, 8, 16 and 20.
Fixing these sizes allows to write dedicated ``kernels'', \emph{i.e.}, sequences of instructions without explicit loops or tests, that aim to maximize arithmetic throughput.

When generic scalar CPU optimizations are exhausted, Faiss also optimizes specifically for some hardware platforms. 

\myparagraph{CPU vectorization} \label{app:FastScan}
Modern CPUs support Single Instruction, Multiple Data (SIMD) operations, specifically AVX/AVX2/AVX512 for x86 and NEON for ARM. 
Faiss exploits those at three levels. 

When operations are simple enough (\eg elementwise vector sum), the code is written in a way that the compiler can vectorize the code by itself, which often boils down to adding \verb|restrict| keywords to promise that arrays are not overlapping. 

The second level leverages SIMD variables and instructions through  C++ compiler extensions. 
Faiss includes \faissfunction{simdlib}, a collection of classes intended as a layer above the AVX and NEON instruction sets. 
However, much of the SIMD is done specifically for one instruction set – most often AVX – because it is more efficient. 

The third level of optimization adapts the data layout and algorithms in order to speed up their SIMD implementation.
The 4-bit product and additive quantizer implementations are implemented in this way, inspired by the SCANN library~\cite{guo2020accelerating}: the layout of the PQ codes for several consecutive vectors is interleaved in memory so that a vector permutation can be used to perform the LUT lookups of \eqref{eq:lutlookup} in parallel. 
This is implemented in the \faissfunction{FastScan} variants of PQ and AQ indexes (\faissfunction{IndexPQFastScan}, \faissfunction{IndexIVFResidual} \faissfunction{QuantizerFastScan}, etc.).

\myparagraph{GPU Faiss}
\label{app:gpu}
Porting Faiss to the GPU is an involved undertaking due to substantial architectural specificities. The implementation of GPU Faiss is detailed in \cite{johnson2019billion}, we summarize the GPU implementation challenges therein.

Modern multi-core CPUs are highly latency optimized: they employ an extensive cache hierarchy, branch prediction, speculative execution and out-of-order code execution to improve serial program execution. In contrast, GPUs have a limited cache hierarchy and omit many of these latency optimizations. They instead possess a larger number of concurrent threads of execution (Nvidia's A100 GPU allows for up to 6,912 \textit{warps}, each roughly equivalent to a 32-wide vector SIMD CPU thread of execution), a large number of floating-point and integer arithmetic functional units (A100 has up to 19.5 teraflops per second of fp32 fused-multiply add throughput), and a massive register set to allow for a high number of long latency pending instructions in flight (A100 has 27 MiB of register memory). They are thus largely throughput-optimized machines.

The algorithmic techniques used in vector search can be grouped into three broad categories: distance computation of floating-point or binary vectors (which may have been produced via dequantization from a compressed form), table lookups (as seen in PQ distance computations) or scanning (as seen when traversing IVF lists), and irregular, sequential computations such as linked-list traversal (as used in graph-based indices) or ranking the $k$ closest vectors.

Distance computation is easy on GPUs and readily exceeds CPU performance, as GPUs are optimized for matrix-matrix multiplication such as that seen in \faissfunction{IndexFlat} or \faissfunction{IVFFlat}. Table lookups and list scanning can also be made performant on GPUs, as it is possible to stage small tables (as seen in product quantization) in \textit{shared memory} (roughly a user-controlled L1 cache) or register memory and perform lookups in parallel across all warps. 

Sequential table scanning in IVF indices requires loading data from main (\textit{global}) memory. While main memory access latency is high, for table scanning we know in advance what data we wish to access. The data movement from main memory into registers can thus be pipelined or use double buffering, so we can achieve close to peak possible performance.

Selecting the $k$ closest vectors to a query vector by ranking distances on the CPU is best implemented with a min- or max-heap. On the GPU, the sequential operations involved in heap operations would similarly force the GPU into a latency-bound regime. This is the largest challenge for GPU implementation of vector search, as the time needed for the heap implementation is an order of magnitude greater than all other arithmetic. To handle this, we developed an efficient GPU $k$-selection algorithm~\cite{johnson2019billion} that allows for ranking candidate vectors in a single pass, operating at a substantial fraction of peak possible performance per memory bandwidth limits. It relies upon heavy usage of the high-speed, large register memory on GPUs, and small-set bitonic sorting via \textit{warp shuffles} with buffering techniques.

Irregular computations such as walking graph structures for graph-based indices like HNSW tend to remain in the latency-bound (due to the sequential traversal) rather than arithmetic throughput or memory bandwidth-bound regimes. Here, GPUs are at a disadvantage as compared to CPUs, and emerging techniques such as CAGRA \cite{ootomo2023cagra} are required to parallelize otherwise sequential operations with graph traversal. 

GPU Faiss implements brute-force \faissfunction{GpuIndexFlat} as well as the IVF indices \faissfunction{GpuIndexIVFFlat}, \faissfunction{GpuIndexIVF} \faissfunction{ScalarQuantizer} and \faissfunction{GpuIndexIVFPQ}, which are the most useful for large-scale indexing. The coarse quantizer for the IVF indices can be on either CPU or GPU.
The GPU index objects have the same interface as their CPU counterparts and the functions \faissfunction{index\_cpu\_to\_gpu} / \faissfunction{index\_gpu\_to\_cpu} convert between them. 
Multiple GPUs are also supported. 
GPU indexes can take inputs and outputs in GPU or CPU memory as input and output, and Python interface can handle Pytorch tensors.

\myparagraph{Advanced options for Faiss components and indices}
Many Faiss components expose internal parameters to fine-tune the trade-off between metrics: 
number of iterations of k-means, 
batch sizes for brute-force distance computations, etc. 
Default parameter values are set to work reasonably well in most cases.

\myparagraph{Multi-threading}
Faiss relies on OpenMP to handle multi-threading. 
By default, Faiss switches to multi-threading processing if it is beneficial, for example, at training and batch addition time. 
Faiss multi-threading behavior may be controlled with standard OpenMP environment variables and functions, such as \faissfunction{omp\_set\_num\_threads}.

When searching a single vector, Faiss does not spawn multiple threads. 
However, when batched queries are provided, Faiss processes them in parallel, exploiting the effectiveness of the CPU cache and batched linear algebra operations. 
This is faster than calling \faissfunction{search} from multiple threads. 
Therefore, queries should be submitted by batches if possible.

\subsection{Interfacing with external storage}
\label{app:externalstorage}

Faiss indexes are based on simple storage structures, mainly \verb|std::vector| to make copy-construction easier. 
The default implementation of \faissfunction{IndexIVF} is based on this storage. 
However, to give vector database developers more control over the storage of inverted lists, Faiss provides two lower-level APIs.

\myparagraph{Arbitrary inverted lists}
The IVF index uses an abstract \faissfunction{InvertedLists} object as its storage.
The object exposes routines to read one inverted list, add entries to it and remove entries.
The default \faissfunction{ArrayInvertedLists} uses in-memory storage. 
Alternatively,  \faissfunction{OnDiskInverted} \faissfunction{Lists} provides memory-mapped storage. 

More complex implementations can access a key-value storage either by storing the entire inverted list as a value, or by utilizing key prefix scan operations like the one supported by RocksDB to treat multiple keys prefixed by the same identifier as one inverted list. 
To this end, the \faissfunction{InvertedLists} implementation exposes an \faissfunction{InvertedListsIterator} and fetches the codes and ids from the underlying key-value store, which usually exposes a similar iterable interface. 
Adding, updating and removing codes can be delegated to the underlying key-value store. We provide an implementation for RocksDB in \faissfunction{rocksdb\_ivf}.

\myparagraph{Scanner objects}
With the abstraction above, the scanning loop is still controlled by Faiss. 
If the calling code needs to control the looping code, then the Faiss IVF index provides an \faissfunction{InvertedListScanner} object. 
The scanner's state includes the current query vector and current inverted list. 
It provides a \faissfunction{distance\_to\_code} method that, given a code, computes the distance from the query to the decompressed vector. 
At a slightly higher level, it loops over a set of codes and updates a provided result buffer. 

This abstraction is useful when the inverted lists are not stored sequentially or fragmented into sub-lists because of metadata filtering~\cite{huang2020embedding}.
Faiss is used only to perform the coarse quantization and the vector encoding.

\subsection{How to choose an index}
\label{app:decisiontree}

\begin{figure}[t]
    \centering
    \includegraphics[width=\linewidth]{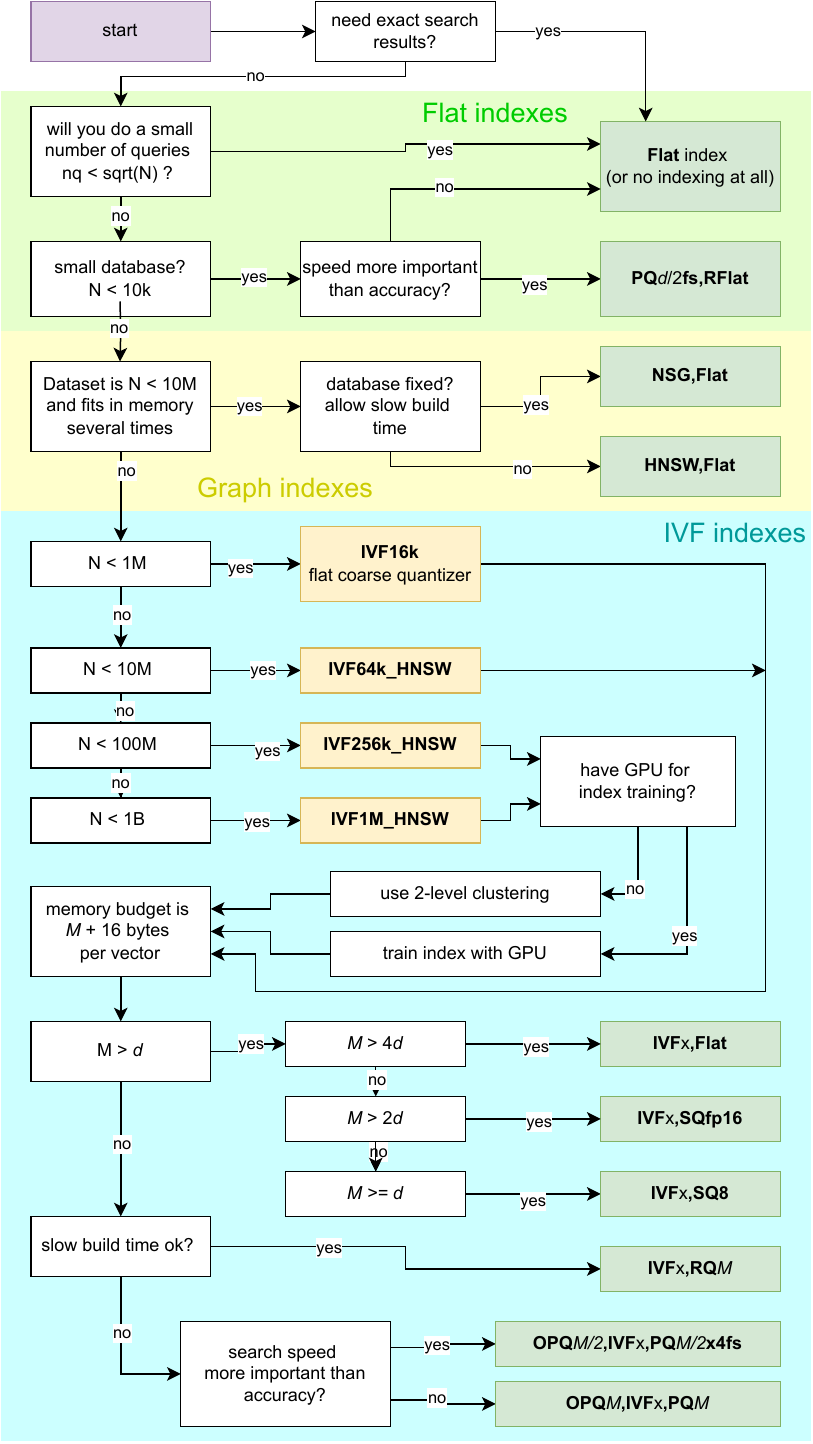}
    \caption{Decision tree to choose a Faiss index.
    This is for the common case of Euclidean k-nearest neighbor search on CPU.
    The resulting indexes (in bold) are defined by their factory string (see Appendix~\ref{app:highlevel}).
    }
    \label{fig:decisiontree}
\end{figure}

Figure~\ref{fig:decisiontree} shows a decision tree for index types, depending on database size and memory constraints. 
The decision tree first focuses on the ``hard'' \metric{memory} constraint, then on secondary trade-offs, like \metric{index construction time} vs. \metric{accuracy}.

\subsection{Faiss metric types}
\label{app:metrics}

Table~\ref{tab:metrics} lists the metric types supported in Faiss. 
They are inspired by the distance metrics in the Scipy library~\cite{2020SciPy-NMeth}.
Only the two first metric types are supported for all index types. 
The others are supported only for brute-force computations and non-compressed versions of IVF and HNSW (\faissfunction{IVFFlat} and \faissfunction{HNSWFlat}).

\begin{table*}[]
    \centering    
\caption{
Faiss metric types and corresponding function in \texttt{scipy.distance}. Note that the Faiss names are often more terse than their scipy counterparts. 
}
    \begin{tabular}{l|llp{50mm}}
    \toprule
        Faiss name & scipy name & definition & comments \\
        \midrule
\faissfunction{METRIC\_INNER\_PRODUCT} &  & $\langle x, y \rangle$ &  \\
\faissfunction{METRIC\_L2}            & sqeuclidean & $\|x-y\|^2$ & The Eucliean metric can be obtained via a square root \\
\faissfunction{METRIC\_L1}            &  cityblock & $\sum_i |x_i - y_i|$ &  \\
\faissfunction{METRIC\_Linf}          & chebyshev & $\max_i |x_i - y_i|$ &  \\
\faissfunction{METRIC\_Lp}            & minkowski & $\sum_i |x_i-y_i|^p$ & the parameter $p$ is set via the \faissfunction{metric\_arg} parameter\\
\faissfunction{METRIC\_Canberra}      & canberra &  $\sum_i |x_i-y_i| / (|x_i| + |y_i|)$ & \\
\faissfunction{METRIC\_BrayCurtis}    & braycurtis & $\sum_i |x_i-y_i| / \sum_i(|x_i + y_i|)$ & \\
\faissfunction{METRIC\_JensenShannon} & jensenshannon & $\frac{1}{2} (\mathrm{KL}(x || m) + \mathrm{KL}(y || m))$ & where $m=\frac{1}{2}(x + y)$ and $\mathrm{KL}$ is the Kullback-Leiber divergence between two distributions\\
\faissfunction{METRIC\_Jaccard} & jaccard & $\sum_i \min(x_i, y_i) / \sum_i \max(x_i, y_i)$ &  \\
\faissfunction{METRIC\_NaNEuclidean} & & $\frac{|V|}{d} \sum_{i\in V} (x_i - y_i)^2$ & $V = \{i\in\{1..d\} \, \, \mathrm{ st.}\,\, x_i \ne \mathrm{NaN} \wedge x_i \ne \mathrm{NaN} \} $\\
\bottomrule
    \end{tabular}
    \label{tab:metrics}
\end{table*}

\subsection{API index of the Faiss library}

As mentioned in Section~\ref{sec:howtochoseindex}, most indexes are a combination of a compression method (like PQ) and a non-exhaustive search method. 
The index class names are built as shown in Table~\ref{tab:combinations}. 

\begin{figure*}
    \centering
    \includegraphics[width=\linewidth]{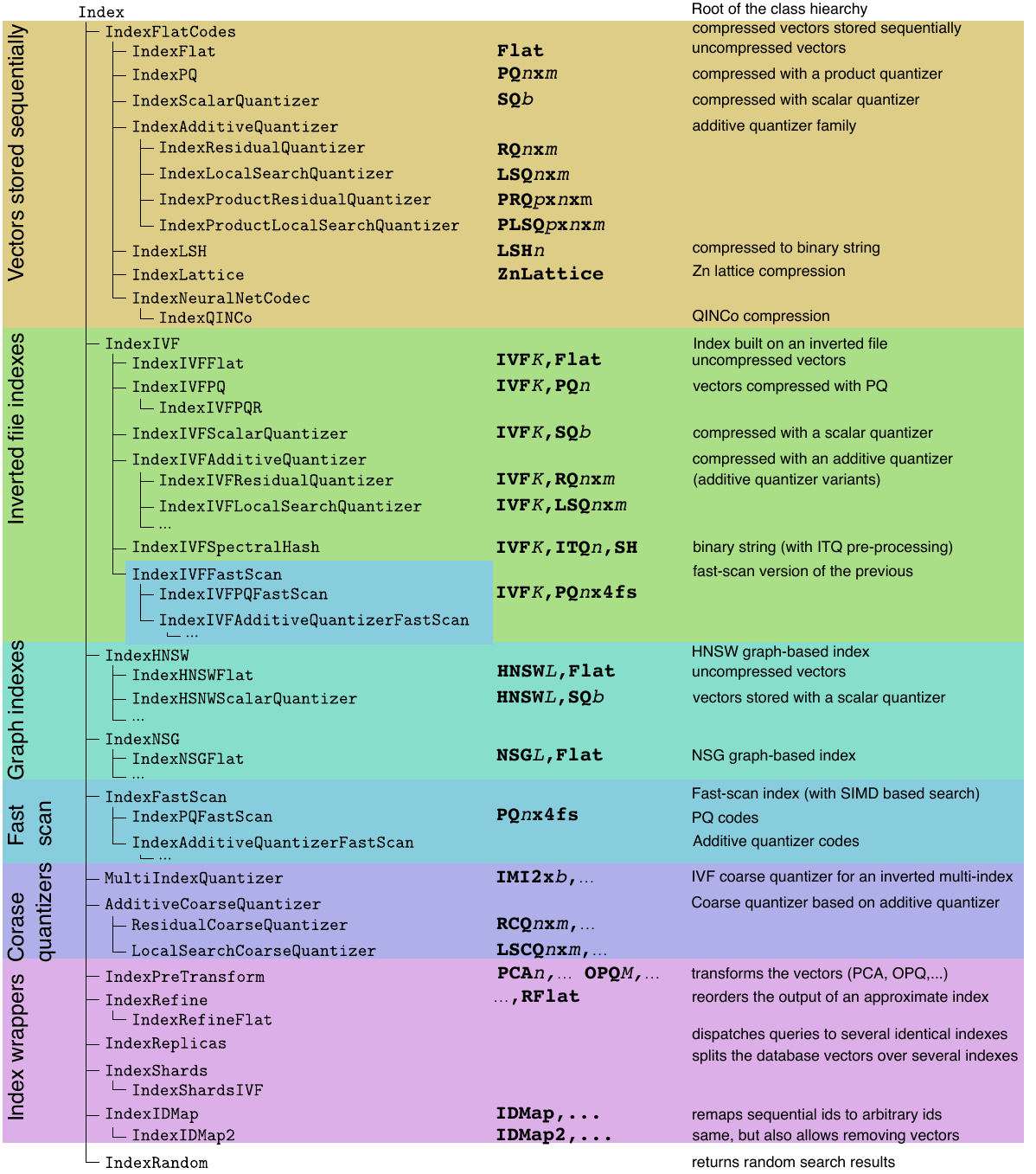}
    \caption{ Hierarchy of the main index classes in CPU Faiss for floating-point vectors.
    For each class we indicate the corresponding factory string (when applicable) and a short explanation.
    }
    \label{fig:classhierachy}
\end{figure*}

Figure~\ref{fig:classhierachy} shows the most important index classes in Faiss, grouped by broad families. 
Some combinations are omitted for brevity.

\begin{table}[]
    \caption{A few combinations of pruning approaches (rows) and compression methods (columns). 
    In the cells: the corresponding index implementations.}
    \label{tab:combinations}
    \centering
\hspace*{-5mm}
    \begin{tabular}{@{\ \ }c|l@{\ \ \ }l@{\ \ \ }l@{\ }}
    \toprule
        & No encoding & PQ encoding & scalar quantizer \\
    \midrule
        Flat &  \faissfunction{IndexFlat} & \faissfunction{IndexPQ} & \faissfunction{IndexScalarQuantizer} \\
        IVF & \faissfunction{IndexIVFFlat} & \faissfunction{IndexIVFPQ} & \faissfunction{IndexIVFScalarQuantizer}\\
        HNSW & \faissfunction{IndexHSNWFlat} & \faissfunction{IndexHNSWPQ} & \faissfunction{IndexHNSWScalarQuantizer}\\
    \bottomrule
    \end{tabular}
\end{table}

\myparagraph{Faiss index API}
In the following, we list the main methods offered by Faiss indexes,  $x\in \mathbb{R}^{n\times d}$ is a list of $n$ vectors in dimension $d$ represented as the rows of a matrix, and $I\in\{0, 2^{63}-1\}^n$ is a list of 63-bit ids of size $n$. 
\begin{itemize}
\item \faissfunction{train($x$)} 
    perform a training using vectors $x$ to prepare adding vectors of the same data distribution to the index;
\item \faissfunction{add($x$)} 
    add the vectors $x$ to the index, numbered sequentially; 
\item \faissfunction{add\_with\_ids($x$,$I$)} 
    add the vectors $x$, identified by the 63-bit ids $I$; 
\item \faissfunction{search($x$, $k$)} 
    return the $k$ nearest vectors of each of the query vectors in $x$;
\item \faissfunction{range\_search($x$, $\varepsilon$)} 
    return all vectors within a radius $\varepsilon$ of each of the query vectors in $x$; 
\item \faissfunction{remove\_ids($I$)} 
    remove vectors with ids $I$ from the index; 
\item \faissfunction{reconstruct\_batch($I$)} 
    extract the vectors with ids in $I$. It is called ``reconstruct'' because for most index types, the returned vectors will be approximations of the original ones.
\end{itemize}

\myparagraph{Quantizer objects}
\label{app:quantizers}
Table~\ref{tab:quantizernesting} shows the hierarchy of quantizers. 
Each quantizer can represent the reproduction values of all the quantizers below, but is slower to train and to perform assignment with. 

The root \faissfunction{Quantizer} class has the following fields and methods (where $x\in \mathbb{R}^{n\times d}$ is a list of vectors): 
\begin{itemize}
\item \faissfunction{code\_size} 
    size  of the codes it produces (in bytes); 
\item \faissfunction{train($x$)} 
    perform a training using vectors $x$ to prepare the quantizer;
\item \faissfunction{compute\_codes($x$)} 
    encode a $n$ vectors to an array of size $n\times$(\faissfunction{code\_size}); 
\item \faissfunction{decode($C$)} 
    decode the vectors from codes obtained with \faissfunction{encoder};   
\end{itemize}

\begin{table}[]
    \caption{
    The hierarchy of quantizers. 
    Each quantizer can represent the set of reproduction values of the quantizers below it. 
    }
    \centering
    \begin{tabular}{l|l}
        \toprule
        Type of quantizer          & class \\
        \midrule
        Vector quantizer           & \\
        Additive quantizer         & \faissfunction{AdditiveQuantizer} \\
        Product-additive quantizer & \faissfunction{ProductAdditiveQuantizer} \\
        Product quantizer          & \faissfunction{ProductQuantizer} \\
        Scalar quantizer           & \faissfunction{ScalarQuantizer} \\
        Binarization               & \\
    \bottomrule
    \end{tabular}
    \label{tab:quantizernesting}
\end{table}

\end{document}